\documentclass[11pt]{article}

\usepackage[]{acl}

\usepackage{times}
\usepackage{latexsym}

\usepackage[T1]{fontenc}
\usepackage[utf8]{inputenc}
\usepackage{microtype}
\usepackage{inconsolata}
\usepackage{amsmath}
\usepackage{threeparttable}
 
\usepackage{graphicx}
\usepackage{amsmath}
\usepackage{hyperref}
\usepackage[table]{xcolor}
\usepackage{amsfonts}
\usepackage{amssymb}
\usepackage{pifont}
\usepackage{booktabs}
\usepackage{enumitem}
\usepackage{xcolor}
\usepackage{url}
\usepackage[most]{tcolorbox}
\newcommand{\cmark}{\ding{51}}
\newcommand{\xmark}{\ding{55}}
\definecolor{clinblue}{RGB}{138,110,172}
\definecolor{clinlight}{RGB}{241,239,246}
\definecolor{clinband}{RGB}{223,234,246}

\newcommand{\clinhead}[1]{\textcolor{white}{\textbf{#1}}}
\newcommand{\clinmetric}[1]{\textbf{#1}}

\definecolor{clindeep}{RGB}{92,64,128}
\newtcolorbox{clinbox}[1]{breakable, enhanced, colback=clinlight,
  colframe=clinblue, boxrule=0.7pt, arc=2.5pt, left=6pt, right=6pt,
  top=4pt, bottom=4pt, title={#1}, fonttitle=\bfseries\small,
  coltitle=white, colbacktitle=clinblue, attach boxed title to top left=
  {yshift=-2mm, xshift=4mm}, boxed title style={colframe=clinblue,
  arc=2pt, boxrule=0pt}}
\newtcolorbox{clinstage}[1]{breakable, enhanced, colback=white, colframe=clindeep,
  boxrule=0.6pt, arc=2pt, left=6pt, right=6pt, top=3pt, bottom=3pt,
  fonttitle=\bfseries\footnotesize, coltitle=clindeep, colbacktitle=white,
  title={#1}}

\newcommand{\ours}{\textsc{ClinEnv}}

\title{\ours{}: An Interactive Multi-Stage Long Horizon EHR Environment for Agents}

\author{
Yuxing Lu$^{1,2}$,
Yushuhong Lin$^{1}$,
Wenqi Shi$^{3}$,
J. Ben Tamo$^{1}$\\
\textbf{Xukai Zhao}$^{4}$,
\textbf{Jinzhuo Wang}$^{2}$,
\textbf{May Dongmei Wang}$^{1}$\\[0.75em]
$^{1}$Georgia Institute of Technology\\
$^{2}$Peking University\\
$^{3}$University of Texas Southwestern Medical Center\\
$^{4}$Tsinghua University
}

\begin{document}
\maketitle

\begin{abstract}
Clinical practice is not the selection of an answer from enumerated options: a physician gathers heterogeneous information incrementally and commits to sequential, irreversible decisions under uncertainty. Static benchmarks cannot probe and existing interactive medical benchmarks each compromise on at least one of them. We present \ours{}, an interactive benchmark that evaluates LLMs as attending physicians over real inpatient admissions under a paradigm we term Longitudinal Inpatient Simulation. Each case is automatically constructed into an ordered sequence of decision stages; at every stage the model must actively query four specialized agents before committing to medications, procedures, and diagnoses. \ours{} scores both what the model decides, through deterministic ontology-grounded matching, and how it gathers information. Across seven models, the strongest reaches only 0.31 decision F1, and outcome quality is sharply decoupled from process quality. Difficulty concentrates in management decisions and later stages, where models recover discharge diagnoses far more reliably than management actions (0.51 vs. 0.17 F1) and continue to issue redundant queries as cases progress. \ours{} makes this information-acquisition gap, invisible to outcome-only evaluation, directly measurable.
\end{abstract}

 % We introduce CLINENV, an interactive benchmark that evaluates LLMs as attending physicians over real inpatient admissions under a paradigm we term Longitudinal Inpatient Simulation. CLINENV is built on three principles: (i) automated case construction that converts raw EHR admissions into ordered multi-stage cases without manual annotation, (ii) an interactive multi-agent environment that withholds clinical information until the model actively queries one of four specialized agents (patient, nurse, laboratory, history), and (iii) dual evaluation that scores both decision accuracy through ontology-grounded matching and process quality through coverage and cost metrics. CLINENV spans 3,509 admissions, 9,297 decision stages, and 26,043 ground-truth decisions covering 2,128 distinct ICD codes and 488 drugs. Across seven models, the strongest reaches only 0.31 decision F1, and outcome quality is sharply decoupled from process quality. Difficulty concentrates in management decisions and later stages, where models recover discharge diagnoses far more reliably than management actions (0.51 vs. 0.17 F1) and continue to issue redundant queries as cases progress. CLINENV makes this information-acquisition gap, invisible to outcome-only evaluation, directly measurable.

\section{Introduction}
\label{sec:intro}
Large language models now approach or exceed expert performance on medical licensing examinations~\citep{singhal2023large,nori2023capabilities} and structured clinical question answering~\citep{jin2021disease}. Yet real inpatient practice is not the selection of an answer from enumerated options. An attending physician navigates a hospitalization in which information is incomplete and must be actively sought, decisions are sequential and irreversible, and each decision spans medications, procedures, and diagnoses that are eventually verified
against the patient's recorded course. Static multiple-choice benchmarks cannot probe any of these properties.

Coding agents illustrate how evaluation infrastructure drives progress, moving from static functional tests~\citep{chen2021evaluating} to executable interactive environments~\citep{jimenez2024swe} that double as training signals. Medical evaluation has begun a similar shift, but existing interactive benchmarks each make a structural compromise. Data-interaction benchmarks~\citep{lee2022ehrsql,jiang2025medagentbench} treat the EHR as a queryable database and score atomic operations, measuring data-access skill rather than clinical reasoning. Diagnostic-conversation benchmarks~\citep{schmidgall2024agentclinic,luo2026meddialbench} rely on LLM-simulated patients, bounding ground truth to short exam vignettes and scoring via LLM-as-judge. Coding-style executable benchmarks~\citep{xu2025medagentgym,mao2025medcalc} reduce clinical reasoning to code or formulas the model must produce. None evaluates an LLM as an attending physician managing a real, multi-stage hospitalization with deterministic verification of its decisions.

\begin{figure*}[t]
    \centering
    \includegraphics[width=\linewidth]{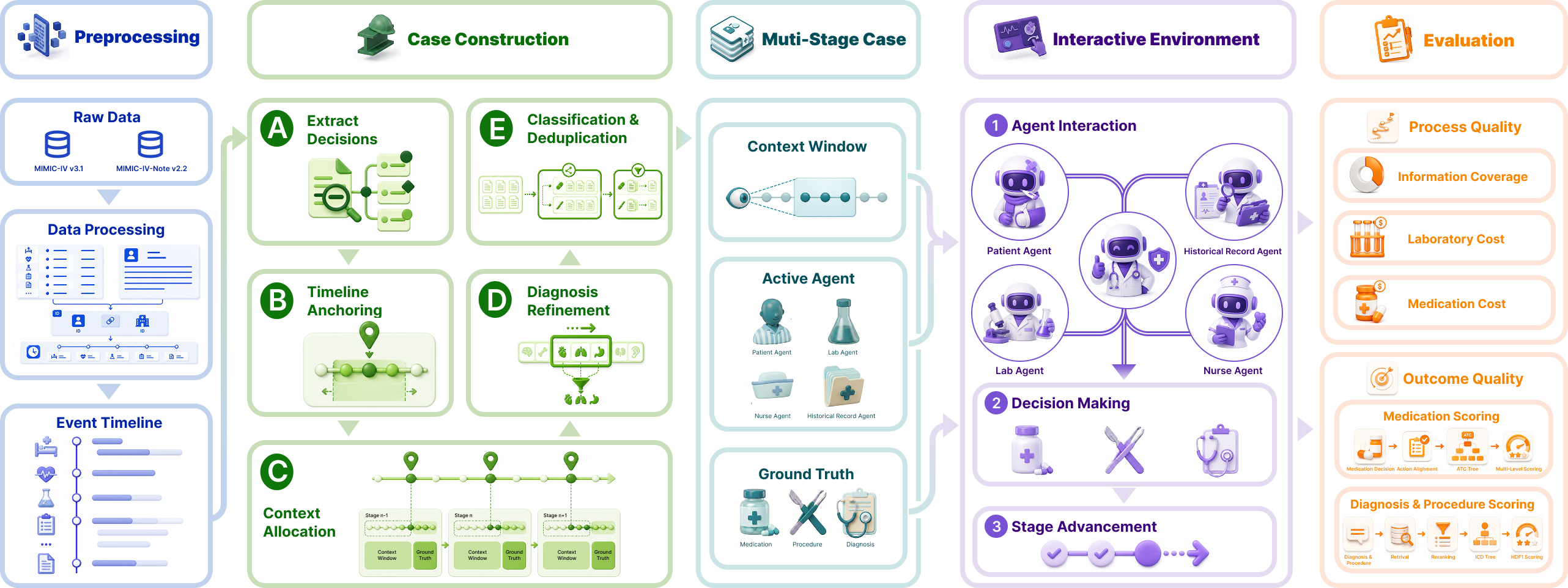}
    \caption{\textbf{Overview of \ours{}.} Patients' admissions are preprocessed into event timelines, converted into multi-stage cases via a five-step pipeline, and evaluated in an interactive environment where the model queries specialized agents before committing decisions, scored on both process and outcome quality.}
    \label{fig:overview}
\end{figure*}

We introduce \ours{}, which evaluates LLMs as attending physicians navigating real admissions through three components (Figure~\ref{fig:overview}). An automated pipeline converts raw admissions into ordered multi-stage cases with structured ground-truth decisions extracted from the EHR timeline and discharge documentation, requiring no manual annotation. An interactive multi-agent environment withholds clinical information until requested: at each stage the model must query four specialized agents (patient, nurse, laboratory, history) before committing to decisions. A dual evaluation framework scores both what the model decides, via deterministic ontology-grounded matching (ATC for medications, hierarchical ICD F1 for diagnoses and procedures), and how it gathers information, via process metrics for coverage and laboratory and medication cost efficiency. We refer to the paradigm defined by these properties, real EHR trajectories, multi-stage decisions, active information acquisition, and deterministic process-aware scoring, as \textbf{Longitudinal Inpatient Simulation}. Across seven LLMs, we find that even the strongest reaches only 0.31 decision F1 and recovers diagnoses far more reliably than management (0.51 vs.\ 0.17 F1), suggesting that outcome-only benchmarks may substantially overestimate clinical readiness and that information seeking is a distinct, measurable dimension of clinical competence. Our contributions are: 

\noindent(i) An automated case construction pipeline that converts raw admissions into multi-stage cases without manual annotation (Section~\ref{sec:planner}); 

\noindent(ii) An interactive multi-agent clinical environment requiring active information acquisition before each decision (Section~\ref{sec:environment},~\ref{sec:benchmark}); 

\noindent(iii) A dual evaluation framework covering decision accuracy and process quality (Section~\ref{sec:evaluation}).

\section{Related Work}
\label{sec:related}

We organize prior medical LLM benchmarks by the capability they evaluate (Table~\ref{tab:related-comparison}).
% more detail is provided in Appendix~\ref{app:related-extended}.

\paragraph{Knowledge benchmarks.} PubMedQA~\citep{jin2019pubmedqa}, MedMCQA~\citep{pal2022medmcqa}, MedQA~\citep{jin2021disease}, the medical subsets of MMLU~\citep{hendrycks2020measuring}, and the aggregated MultiMedQA~\citep{singhal2023large} probe factual recall through specified multiple-choice questions. This family equates clinical competence with knowledge retrieval, with no information acquisition or commitment under uncertainty.

\paragraph{Data-interaction benchmarks.} EHRSQL~\citep{lee2022ehrsql}, MIMIC-SQL~\citep{wang2020text}, and FHIR-AgentBench~\citep{lee2025fhir} treat the EHR as a queryable artifact accessed via SQL or APIs; MedAgentBench~\citep{jiang2025medagentbench} extends this with physician-authored interactive tasks on 100 patient profiles. These benchmarks measure proficiency at executing well-specified operations against patient data. Their tasks are atomic and externally defined, so they assess data-access skill rather than the clinical reasoning that determines which data matters or what action it warrants.
\begin{table}[t]
\centering
\footnotesize
\setlength{\tabcolsep}{1.2pt}
\renewcommand{\arraystretch}{0.85}
\caption{Design-dimension coverage across medical LLM benchmarks.
\cmark{} = supported, $\triangle$ = partial, \xmark{} = not supported.
Extended version in Appendix~\ref{app:related-extended}.}
\vspace{-5pt}
\label{tab:related-comparison}
\begin{tabular}{lccccc}
\toprule
\rowcolor{clinblue}
\clinhead{} & \clinhead{Real} & \clinhead{Multi-} & \clinhead{Active} & \clinhead{Det.} & \clinhead{Proc.} \\
\rowcolor{clinblue}
\clinhead{Benchmark} & \clinhead{EHR} & \clinhead{stage} & \clinhead{info} & \clinhead{score} & \clinhead{metrics} \\
\midrule
MedQA / MultiMedQA   & \xmark      & \xmark      & \xmark      & \cmark      & \xmark \\
\rowcolor{clinlight}
EHRSQL / MIMIC-SQL   & \cmark      & \xmark      & \xmark      & \cmark      & \xmark \\
FHIR-AgentBench      & \cmark      & \xmark      & \xmark      & \cmark      & \xmark \\
\rowcolor{clinlight}
MedAgentBench        & $\triangle$ & \xmark      & \cmark      & \cmark      & \xmark \\
AgentClinic          & \xmark      & \xmark      & \cmark      & \xmark      & \xmark \\
\rowcolor{clinlight}
MedDialBench / MAQuE & \xmark      & \xmark      & \cmark      & \xmark      & \xmark \\
HELPMed / AMIE       & \xmark      & \xmark      & \cmark      & \xmark      & \xmark \\
\rowcolor{clinlight}
MedCalc-Env          & \xmark      & $\triangle$ & \xmark      & \cmark      & \xmark \\
MedAgentGym          & $\triangle$ & \xmark      & \cmark      & \cmark      & \xmark \\
\midrule
\rowcolor{clinlight}
\textbf{\ours{}}     & \textbf{\cmark} & \textbf{\cmark} & \textbf{\cmark} & \textbf{\cmark} & \textbf{\cmark} \\
\bottomrule
\end{tabular}
\vspace{-12pt}
\end{table}

\paragraph{Diagnostic-conversation benchmarks.}
AgentClinic~\citep{schmidgall2024agentclinic} pairs a doctor agent with an LLM-played patient agent and a measurement agent that expose information only on query, and the doctor must converge on a diagnosis. MedDialBench~\citep{luo2026meddialbench}, MAQuE~\citep{gong2025dialogue}, HELPMed~\citep{bean2025clinical}, and AMIE~\citep{tu2024towards} extend this paradigm with adversarial behaviors, behavioral layering, or human participants. This family established the importance of interactive evaluation, but two limitations follow from its simulator-based design: ground truth is bounded by what the simulator was given (typically exam vignettes), and intermediate behavior is scored via LLM-as-judge over free-text dialogue. \ours{} addresses both by deriving multi-decision trajectories from real admission records and scoring each decision deterministically against structured EHR ground truth via Hungarian matching and ontology-based partial credit.

\paragraph{Coding-style executable benchmarks.}
MedAgentGym~\citep{xu2025medagentgym} and MedCalc-Env~\citep{mao2025medcalc} draw an analogy to coding agent benchmarks (HumanEval~\citep{chen2021evaluating}, SWE-bench~\citep{jimenez2024swe}), proposing executable environments for biomedical code and calculation. \ours{} shares this spirit but targets clinical decisions themselves (which drug to start, which diagnosis to commit to), verified against medical ontologies physicians use, rather than code or formulas the model must produce.

\paragraph{The LIS paradigm.}
Table~\ref{tab:related-comparison} confirms that no existing benchmark jointly satisfies the dimensions defining the LIS paradigm (Section~\ref{sec:intro}): real EHR longitudinal trajectories, multi-stage sequential decisions, active information acquisition across heterogeneous sources, deterministic ontology-based scoring, and cost-aware process metrics. \ours{} is the first to instantiate all five.

\section{\ours{} Construction}
\label{sec:pipeline}

\subsection{Data Preprocessing}
\label{sec:data}

\ours{} is built from MIMIC-IV v3.1~\citep{johnson2023mimic} and MIMIC-IV-Note v2.2~\citep{PhysioNet-mimic-iv-note-2.2}. For each hospitalization, we combine structured EHR tables with clinical notes into an admission-level timeline by \texttt{subject\_id} and \texttt{hadm\_id}. Structured sources include admissions, laboratory and microbiology results, medication orders and administrations, ICU charted events, fluid input/output events, procedure events, and ICD-coded diagnoses and procedures; note sources include discharge notes and radiology reports. Each retained row becomes a provenance-bearing event with a canonical \texttt{event\_time}, \texttt{source\_table}, and the original table-specific clinical fields. Events are sorted into one timeline per admission. Notes supply narrative context for decision extraction and history access, while structured records provide timestamped observations and coded ground truth for evaluation. Appendix~\ref{app:data-preprocessing} describes the preprocessing procedure and examples.

\subsection{Automated Case Construction}
\label{sec:planner}
Each admission is processed by a four-phase pipeline that segments the patient timeline into an ordered sequence of clinical stages, with ground-truth decision sets derived entirely from structured records and discharge documentation. Claude-Sonnet-4.6 is used as the planning backbone throughout.

\paragraph{Phase A: Decision Extraction.}
The planner reads the discharge note to extract every attending-level physician decision made during the hospitalization. When a full free-text discharge note is available it is used directly; otherwise, four sections are stitched together as a fallback: Chief Complaint, History of Present Illness, Brief Hospital Course, and Discharge Diagnosis. Each decision is annotated with a type (medication, procedure, or plan).

\paragraph{Phase B: Temporal Anchoring.} 
Given a timeline of $T$ events and decisions $d_1, \ldots, d_n$, Phase B localizes each decision to a specific event or event range via a sequential forward sliding-window agent. Let $\ell_0 = 0$. For decision $d_i$, a window 
\begin{equation}
\scriptsize
\mathcal{W}(s) = \bigl[s,\; \min(s + W - 1,\; T - 1)\bigr]
\end{equation}
is presented to an LLM agent equipped with \texttt{mark\_single} and \texttt{mark\_range} tool calls, beginning at $s = \ell_{i-1}$. If the agent marks anchor $a_i$ within the window, the anchor is recorded and $\ell_i = a_i$; the next decision $d_{i+1}$ begins its search from $\ell_i + 1$. If no anchor is found, the window advances ($s \leftarrow s + W$) and the agent retries. If the timeline is exhausted without a match, $d_i$ is left unanchored and $\ell_i = \ell_{i-1}$. Decisions anchored exclusively to pre-admission events are discarded.

\paragraph{Phase C: Stage Construction and Validation.} Given anchor index $d_i$ for decision $i$, the context window for stage $k$ is defined as $\mathcal{C}_k = [\mathit{prev\_end} + 1,\; d_i - 1]$, where $\mathit{prev\_end}$ is the last event index consumed by the preceding stage. A decision is skipped if (a) no context window exists ($\mathit{prev\_end} \geq d_i - 1$), or (b) the decision is a medication whose drug is already actively present in $\mathcal{C}_k$. A final diagnosis stage is appended to every case: its context spans all events from the last decision through discharge, and its ground truth is derived from a separate diagnostic scan.

\begin{table*}[t]
\centering
\footnotesize
\setlength{\tabcolsep}{3pt}
\renewcommand{\arraystretch}{0.9}
\caption{Clinical information agents and their readviews.}
\label{tab:agents}
\begin{tabular}{p{0.08\textwidth}p{0.44\textwidth}p{0.44\textwidth}}
\toprule
\rowcolor{clinblue}
\clinhead{Agent} & \clinhead{Readview $\mathcal{V}_a$} & \clinhead{Role} \\
\midrule
Patient
& Demographics, chief complaint, history of present illness, past medical history
& Simulates subjective patient reporting in lay language \\

\rowcolor{clinlight}
Nurse
& Vitals, fluid balance, medication administration records
& Reports bedside objective observations \\

Lab
& Laboratory and microbiology results
& Resolves test-name queries and returns test results \\

\rowcolor{clinlight}
History
& Prior discharge summaries
& Supplies longitudinal clinical context \\
\bottomrule
\end{tabular}
\end{table*}

\paragraph{Phase D: Diagnostic Groundability Scan.} 
A sliding-window scan identifies which of the admission's discharge ICD codes are groundable, i.e., actively evidenced or managed within the event timeline. Given a timeline of $T$ events, we partition it into overlapping windows:
\begin{equation}
\scriptsize
\mathcal{W}_i = \bigl[iS,\; \min(iS + W,\, T)\bigr), \quad i = 0, 1, \ldots, \left\lceil \frac{T - W}{S} \right\rceil
\end{equation}
where $W$ is the window size and $S < W$ is the stride. Only ICD codes confirmed by at least one window contribute to the final diagnosis stage.

\paragraph{Phase E: Classification and Enrichment.} Each stage's raw decision description is classified into one of four types (medication, procedure, diagnosis, plan) via LLM. For medication, the action type is further assigned as one of \{\texttt{start}, \texttt{stop}, \texttt{switch}, \texttt{adjust}\}; \texttt{adjust} decisions optionally carry a direction (\texttt{increase} or \texttt{decrease}). Ground-truth fields are then enriched deterministically from EHRs with zero LLM calls: drug names are read from prescription, pharmacy, and eMAR tables; ICD codes for procedures and diagnoses are taken directly from the corresponding coded EHR events. The resulting stage $\mathcal{S}_k$ is defined as:
\begin{equation}
\scriptsize
  \mathcal{S}_k = (\mathcal{C}_k,\; \mathcal{G}_k,\; \mathcal{A}_k)
\end{equation}
where $\mathcal{C}_k = [\mathit{start}_k, \mathit{end}_k]$ is the visible event window, $\mathcal{G}_k = \{g_1^k, \dots, g_{m_k}^k\}$ is the structured ground-truth decision set, and $\mathcal{A}_k \subseteq \{\texttt{patient}, \texttt{nurse}, \texttt{lab}, \texttt{history}\}$ is the set of activated information sources for this stage.

\subsection{Multi-Source Clinical Environment}
\label{sec:environment}
The evaluation environment exposes clinical information through four specialized agents (Table~\ref{tab:agents}), each powered by GPT-5.4-mini and restricted to a designated readview $\mathcal{V}_a$, which is a filtered projection of the EHR events relevant to that role.

The patient agent answers in lay language from demographics and clinical history; the nurse agent reports bedside vital signs, fluid balance, and medication administration; the lab agent resolves natural-language test queries against the stage's available results via semantic matching, returning nothing for tests not on record; and the history agent supplies prior discharge summaries when earlier admissions exist. Table~\ref{tab:agents} lists each readview, and Appendix~\ref{app:agent-details} gives the full agent specifications.

\subsection{Interaction Modes}
\label{sec:modes}

The evaluation interface exposes the information-gathering and
decision-submission tools summarized in Table~\ref{tab:tools}.

\paragraph{Direct Mode.}
The model to be evaluated receives the complete observation for stage $k$:
\begin{equation}
\scriptsize
  \mathcal{O}_k^{\text{direct}} =
  \bigl\{\,\mathcal{P},\;
  \mathcal{E}_{[\mathit{start}_k,\, \mathit{end}_k]}\,\bigr\}
\end{equation}
where $\mathcal{P}$ denotes static patient context and $\mathcal{E}_{[\mathit{start}_k,\mathit{end}_k]}$ is the cumulative event sequence up to the end of the stage. The model submits decisions directly without gathering information.

\begin{table}[t]
\centering
\footnotesize
\setlength{\tabcolsep}{2.5pt}
\renewcommand{\arraystretch}{0.9}
\caption{Available tools by evaluation mode.}
\label{tab:tools}
\begin{tabular}{@{}p{0.38\columnwidth}p{0.17\columnwidth}p{0.38\columnwidth}@{}}
\toprule
\rowcolor{clinblue}
\clinhead{Tool} & \clinhead{Mode} & \clinhead{Function} \\
\midrule
\texttt{ask\_patient}           & Interactive & Query patient agent \\
\rowcolor{clinlight}
\texttt{ask\_nurse}             & Interactive & Query nurse agent \\
\texttt{order\_lab}             & Interactive & Query lab agent \\
\rowcolor{clinlight}
\texttt{get\_history\_summary}  & Interactive & List prior admissions \\
\texttt{get\_history\_detail}   & Interactive & Retrieve full discharge note \\
\midrule
\rowcolor{clinlight}
\texttt{submit\_medication}     & Both & Submit medication decision \\
\texttt{submit\_diagnosis}      & Both & Submit diagnosis \\
\rowcolor{clinlight}
\texttt{submit\_procedure}      & Both & Submit procedure decision \\
\texttt{submit\_plan}           & Both & Submit other management decision \\
\rowcolor{clinlight}
\texttt{finalize\_decision}     & Both & Advance to next stage \\
\bottomrule
\end{tabular}
\end{table}

\paragraph{Interactive Mode.}
The model begins each stage with no clinical data. Its observation at turn $t$ accumulates through sequential tool calls:
\begin{equation}
\scriptsize
  \mathcal{O}_k^{\text{interactive}}(t) =
  \{\,(q_1, r_1),\; (q_2, r_2),\; \ldots,\; (q_t, r_t)\,\}
\end{equation}
where $q_i$ is the $i$-th tool call and $r_i$ the corresponding agent response. Exactly one tool call is permitted per turn. The episode terminates when the model calls \texttt{finalize\_decision} or after 60 turns, whichever comes first.

\paragraph{Structured Submission Interface.}
To isolate clinical reasoning quality from output-format confounds, the submission interface is constrained in two coupled ways. First, only the submit tools corresponding to ground truth types present in $\mathcal{S}_k$ are exposed; types absent from $\mathcal{G}_k$ are withheld entirely. Second, the model receives the exact required submission count per type:
\begin{equation}
\scriptsize
  n_k^{(t)} = \bigl|\{g \in \mathcal{G}_k : g.\mathrm{type} = t\}\bigr|,
  \quad t \in \mathcal{T}_k
\end{equation}
where $\mathcal{T}_k$ denotes the set of ground truth types in stage $\mathcal{S}_k$. Because the model submits exactly $n_k^{(t)}$ items per type, the per-type Hungarian matching operates on equal-sized sets, so precision and recall differ only by match quality rather than count mismatch. Benchmark scores therefore reflect the accuracy of clinical decisions rather than the model's ability to infer the submission format or the number of items expected.

\section{\ours{} Benchmark}
\label{sec:benchmark}
The \ours{} benchmark is publicly available on Physionet,\footnote{\href{https://huggingface.co/datasets/ylin766gatech/ClinEnv}{\ours{} Huggingface Link}. Must meet PhysioNet's compliance standards before use.} and the source code repository is accessible on GitHub.\footnote{\href{https://github.com/ylin766/ClinEnv}{\ours{} GitHub Link}.} We also release a demo website for interactive evaluation.\footnote{\href{https://clinenv-demo.vercel.app/}{\ours{} Interactive Evaluation Demo Link}} We sample 3{,}509 admissions from 1{,}809 patients, keeping only admissions that produce at least one valid decision stage and stratifying by horizon so that short and long admissions are both well represented. This gives 9{,}297 decision stages and 26{,}043 ground-truth decisions (Table~\ref{tab:benchmark-stats}), with 2.65 stages per case and 49.8\% of cases long-horizon ($\geq$3 stages). Diagnoses account for most decisions (71.7\%), followed by medications (21.4\%) and procedures (6.9\%); each medication carries an action type (\texttt{start}, \texttt{stop}, \texttt{switch}, \texttt{adjust}) for the action-gated scoring of Section~\ref{sec:outcome}. Because the cases come from real admissions rather than a curated question bank, \ours{} spans the breadth of inpatient practice: 2{,}128 distinct ICD codes and 488 distinct drugs, with 3.67 information agents available per stage and 32.8 key clinical facts per readview on average. Appendix~\ref{app:benchmark-stats} gives the sampling protocol and construction-quality checks; Figures~\ref{fig:entity-profile} and~\ref{fig:span-density} further visualize the clinical entity distribution and case-length structure.

\begin{table}[t]
\centering
\footnotesize
\setlength{\tabcolsep}{12pt}
\renewcommand{\arraystretch}{0.9}
\caption{\ours{} benchmark composition.}
\label{tab:benchmark-stats}
\begin{tabular}{lr}
\toprule
\rowcolor{clinblue}
\clinhead{Property} & \clinhead{Value} \\
\midrule
Admissions            & 3{,}509 \\
\rowcolor{clinlight}
Unique patients               & 1{,}809 \\
Cases                         & 3{,}509 \\
\rowcolor{clinlight}
\quad single-stage            & 718 (20.5\%) \\
\quad two-stage               & 1{,}043 (29.7\%) \\
\rowcolor{clinlight}
\quad long-horizon ($\geq$3)  & 1{,}748 (49.8\%) \\
Decision stages (total)       & 9{,}297 \\
\rowcolor{clinlight}
Decisions (total)             & 26{,}043 \\
\quad diagnosis               & 18{,}664 (71.7\%) \\
\rowcolor{clinlight}
\quad medication              & 5{,}571 (21.4\%) \\
\quad procedure               & 1{,}808 (6.9\%) \\
\midrule
Medication action types       & \\
\rowcolor{clinlight}
\quad start / stop            & 79.1\% / 7.5\% \\
\quad switch / adjust         & 3.8\% / 9.5\% \\
\midrule
Distinct ICD codes            & 2{,}128 \\
\rowcolor{clinlight}
Distinct drugs                & 488 \\
\midrule
Avg.\ stages per case         & 2.65 \\
\rowcolor{clinlight}
Avg.\ decisions per stage     & 2.80 \\
Avg.\ available agents / stage & 3.67 \\
\rowcolor{clinlight}
Avg.\ key facts per readview  & 32.8 \\
\bottomrule
\end{tabular}
\end{table}

\section{\ours{} Evaluation}
\label{sec:evaluation}

\subsection{Clinical Decision Accuracy}
\label{sec:outcome}
Matching is performed independently within each type group (diagnosis, medication, procedure) via the Hungarian algorithm~\citep{kuhn1955hungarian}, which maximises the total matched score within each group. The results are then concatenated to form the full stage-level assignment. Stage-level F1 is the harmonic mean of continuous precision and recall derived from the matched scores.

\paragraph{Medication Scoring.}
Action type serves as a hard gate (\texttt{start}/\texttt{stop}/\texttt{switch}/\texttt{adjust}) within the medication group: a submission whose action disagrees with its matched ground truth scores zero regardless of drug similarity, driving the Hungarian algorithm to prefer action-consistent assignments. 

When actions agree, drug similarity is scored via ATC-hierarchy partial credit~\citep{chen2012predicting}. An exact name match yields 1.0 directly. Otherwise, both names are resolved to ATC codes via the RxNorm API. Let $\hat{c}$ and $c^*$ denote the predicted and ground-truth ATC codes, and $\ell$ the length of their longest common prefix.
\begin{equation}
\scriptsize
s_{ATC}=
\begin{cases}
    1.0 & \ell \geq |c^*| \\
    0.8 & \ell = 5 \text{ (chemical subgroup)} \\
    0.6 & \ell = 4 \text{ (pharmacological class)} \\
    0.3 & \ell = 3 \text{ (therapeutic group)} \\
    0.1 & \ell = 1 \text{ (anatomical group)} \\
    0.0 & \text{otherwise}
\end{cases}
\end{equation}

The condition $\ell \geq |c^*|$ grants full credit when the predicted code is at least as specific as the ground truth, accommodating drugs whose RxNorm entry does not reach Level 5. For \texttt{adjust} decisions, the final medication score incorporates a direction penalty:
\begin{equation}
\scriptsize
s_{med} = 
\begin{cases}
    s_{ATC} & \text{if direction matches or is unspecified} \\
    0.5 \cdot s_{ATC} & \text{if directions conflict}
\end{cases}
\end{equation}
where a direction conflict arises only when both the submission and the ground truth explicitly specify opposing directions (increase vs.\ decrease); if either omits it, no penalty applies.

\paragraph{Diagnosis and Procedure Scoring.}
Because the model submits free text while ground truth carries structured ICD codes, submitted text is first mapped to a candidate ICD code via a two-step pipeline: (1) an embedding index over the full ICD vocabulary (ICD-9-CM, ICD-10-CM, ICD-9-PCS, ICD-10-PCS) retrieves the top-15 candidates by cosine similarity; (2) an LLM reranker selects the best match. The predicted code $\hat{c}$ and ground-truth code $c^*$ are each expanded to their full ancestor sets in the ICD hierarchy, and Hierarchical F1 (HDF1) is computed over these sets:
\begin{equation}
\scriptsize
  \text{HDF1}(\hat{c}, c^*) =
  \frac{2\,|\hat{\mathcal{A}} \cap \mathcal{A}^*|}
       {|\hat{\mathcal{A}}| + |\mathcal{A}^*|}
\end{equation}
where $\hat{\mathcal{A}} = \{\hat{c}\} \cup \text{anc}(\hat{c})$ and $\mathcal{A}^* = \{c^*\} \cup \text{anc}(c^*)$.

\begin{table*}[t]
\centering
\footnotesize
\setlength{\tabcolsep}{4.5pt}
\renewcommand{\arraystretch}{1}
\begin{threeparttable}
\caption{\textbf{Overall and long-horizon \ours{} performance.} The long-horizon subset contains admissions with at least three decision stages. Values are stage-macro averages. Dx/Rx/Proc are type-specific match scores. Lab waste is the ratio of aggregate wasted lab cost.}
\vspace{-5pt}
\label{tab:main-results-long-horizon}
\begin{tabular}{lccccccccccc}
\toprule
\rowcolor{clinblue}
\clinhead{Model} & \clinhead{\# Stage}
& \multicolumn{4}{c}{\clinhead{Outcome quality $\uparrow$}}
& \multicolumn{3}{c}{\clinhead{Information seeking}}
& \multicolumn{3}{c}{\clinhead{Cost $\downarrow$}} \\
\cmidrule(lr){3-6}\cmidrule(lr){7-9}\cmidrule(lr){10-12}
\rowcolor{clinlight}
& & \textbf{F1} & \textbf{Dx} & \textbf{Rx} & \textbf{Proc.}
& \textbf{Coverage} & \textbf{Efficiency} & \textbf{Turns}
& \textbf{Waste} & \textbf{Lab \$} & \textbf{Rx \$/day} \\
\midrule

\textbf{GPT-5.4}\tnote{*} & All & \clinmetric{0.306} & \clinmetric{0.511} & 0.097 & \clinmetric{0.379} & \clinmetric{0.430} & 0.274 & 9.0 & 10.4\% & 117 & 64 \\
\rowcolor{clinlight}
\quad \textit{Long Horizon} & $\geq$3 & \clinmetric{0.235} & \clinmetric{0.507} & 0.089 & \clinmetric{0.296} & \clinmetric{0.416} & 0.221 & 11.9 & \clinmetric{2.3\%} & 124 & 87 \\

\textbf{GPT-5.4-mini}\tnote{*} & All & 0.258 & 0.448 & 0.083 & 0.295 & 0.214 & 0.152 & 6.8 & 7.0\% & 106 & 72 \\
\rowcolor{clinlight}
\quad \textit{Long Horizon} & $\geq$3 & 0.213 & 0.482 & 0.071 & 0.281 & 0.246 & 0.154 & 9.0 & 6.6\% & 56 & 109 \\

\textbf{GPT-5.4-nano}\tnote{*} & All & 0.254 & 0.438 & 0.085 & 0.322 & 0.331 & \clinmetric{0.281} & \clinmetric{4.4} & \clinmetric{4.1\%} & 178 & 50 \\
\rowcolor{clinlight}
\quad \textit{Long Horizon} & $\geq$3 & 0.215 & 0.456 & 0.089 & 0.270 & 0.340 & \clinmetric{0.257} & \clinmetric{5.8} & 7.3\% & 156 & 75 \\

\addlinespace[1pt]
\textbf{Llama-3.1-70B} & All & 0.226 & 0.366 & \clinmetric{0.101} & 0.246 & 0.140 & 0.090 & 14.1 & 35.8\% & \clinmetric{27} & 68 \\
\rowcolor{clinlight}
\quad \textit{Long Horizon} & $\geq$3 & 0.194 & 0.386 & \clinmetric{0.103} & 0.212 & 0.156 & 0.085 & 18.1 & 10.2\% & \clinmetric{20} & 51 \\

\textbf{Llama-3.1-8B} & All & 0.202 & 0.362 & 0.082 & 0.249 & 0.205 & 0.134 & 8.3 & 6.5\% & 219 & 53 \\
\rowcolor{clinlight}
\quad \textit{Long Horizon} & $\geq$3 & 0.172 & 0.356 & 0.076 & 0.249 & 0.210 & 0.115 & 10.7 & 3.1\% & 270 & 63 \\

\addlinespace[1pt]
\textbf{Gemma-3-27B} & All & 0.228 & 0.386 & 0.087 & 0.258 & 0.180 & 0.105 & 7.6 & 35.3\% & 132 & \clinmetric{41} \\
\rowcolor{clinlight}
\quad \textit{Long Horizon} & $\geq$3 & 0.195 & 0.415 & 0.081 & 0.251 & 0.201 & 0.103 & 8.8 & 24.9\% & 147 & \clinmetric{47} \\

\textbf{Gemma-3-12B} & All & 0.197 & 0.376 & 0.070 & 0.235 & 0.137 & 0.077 & 8.8 & 13.0\% & 184 & 34 \\
\rowcolor{clinlight}
\quad \textit{Long Horizon} & $\geq$3 & 0.164 & 0.400 & 0.072 & 0.218 & 0.136 & 0.065 & 10.9 & 8.5\% & 218 & 31 \\

\bottomrule
\end{tabular}

\begin{tablenotes}[flushleft]
\footnotesize
\item[*] Proprietary LLMs were accessed under a \textbf{Zero Data Retention Policy} required by PhysioNet.
\end{tablenotes}
\vspace{-10pt}
\end{threeparttable}
\end{table*}

\subsection{Clinical Process Quality}
\label{sec:process}

Beyond what the model decides, we evaluate how it gathers information through two process metrics that are only applicable in interactive mode.

\paragraph{Information Elicitation Quality.} For each activated agent role $a \in \mathcal{A}_k$, a judge is given the complete readview $\mathcal{V}_a$ and the responses that agent $a$ returned during the stage. The judge enumerates the key clinical facts $I_a$ present in $\mathcal{V}_a$ and marks each fact as covered or not based on whether it appears in the dialogue. Per-role coverage is:              
\begin{equation}                                         \scriptsize
  \text{cov}_a = \frac{|\{i \in I_a : \text{covered}(i)\}|}{|I_a|}           
\end{equation}                                                      
Stage-level coverage is the macro-average across activated roles:              
\begin{equation}
\scriptsize
  \text{cov} = \frac{1}{|\mathcal{A}_k|} \sum_{a \in \mathcal{A}_k}          
\text{cov}_a                              
\end{equation}                                                                 
Coverage alone does not penalize over-querying. We additionally report an efficiency score that discounts coverage when the model issues more information-gathering queries than there are sources to consult. Let $K$ be the number of information-gathering queries in the stage and $|\mathcal{A}_k|$ the number of activated agents. An efficient interaction extracts what each source holds in roughly one well-formed query, so we normalize by the average number of queries per source:
\begin{equation}
\scriptsize
\text{eff} = \frac{\text{cov}}{\max\left(1,\; K / |\mathcal{A}_k|\right)}
\end{equation}

When $K \leq |\mathcal{A}_k|$, no penalty applies; beyond that, efficiency is suppressed in proportion to the redundancy of the model's querying. Normalizing by the source count rather than by the total number of available facts makes efficiency a distinct axis from coverage, rewarding models that gather relevant information parsimoniously.

\paragraph{Laboratory Cost Efficiency.}
Each lab test ordered by the model is mapped to a reimbursement fee using the CMS Clinical Laboratory Fee Schedule via embedding retrieval and LLM reranking, following the same two-step pipeline as ICD scoring. Tests ordered beyond those evidenced in the ground-truth clinical picture are treated as unnecessary expenditure; the wasted ratio is defined as:
\begin{equation}
\scriptsize
  r_{\text{waste}} =
  \frac{\sum_{t \in \mathcal{T}_{\text{excess}}} \text{fee}(t)}
       {\sum_{t \in \mathcal{T}_{\text{ordered}}} \text{fee}(t)}
\end{equation}
where $\mathcal{T}_{\text{ordered}}$ is the set of tests the model ordered and $\mathcal{T}_{\text{excess}}$ is the subset not supported by the ground-truth record.

\paragraph{Drug Acquisition Cost.} For each stage, we estimate the financial burden of newly initiated pharmacotherapy. The daily cost is estimated via a two-step pipeline. First, the drug name is resolved to an RxNorm CUI, which is then mapped to an NDC code and a unit price $p$ from the NADAC (National Average Drug Acquisition Cost) database. Second, the RxNorm CUI is mapped to an ATC code, from which the WHO-defined daily dose (DDD) is retrieved. Each component is matched against the WHO ATC/DDD index via embedding retrieval followed by LLM reranking. The estimated daily cost for a single drug is then: $c = p \times \mathrm{DDD}$ and the stage-level metric is the total daily cost $C = \sum_{i} c_i$ across all qualifying submissions.

\section{Results}
\label{sec:results}

\subsection{Overall Performance on \ours{}}
Table~\ref{tab:main-results-long-horizon} reports overall performance. GPT-5.4 leads on every outcome metric, with the highest decision F1 (0.306), diagnosis score (0.511), procedure score (0.379), and information coverage (0.430). Even so, it comes nowhere near saturating the benchmark: \ours{} demands active information gathering and structured commitments across diagnoses, medications, and procedures, and is far harder than single-step clinical recognition.

Outcome and process quality come apart. Llama-3.1-70B posts the best medication score overall (0.101) yet has the lowest coverage (0.140), the most turns (14.1), and the highest laboratory waste (35.8\%). GPT-5.4-nano is the mirror image: competitive accuracy with the fewest turns (4.4) and the lowest waste (4.1\%). It reaches most of GPT-5.4's coverage with half the queries, making it the most information-efficient model (0.281 vs.\ 0.274) even though GPT-5.4 gathers more in absolute terms. Endpoint accuracy alone therefore hides large differences in how models gather and spend evidence.

Long-horizon admissions are uniformly harder: every model scores lower on cases with at least three stages than on the full set, and GPT-5.4 drops from 0.306 to 0.235. The next analyses ask how case horizon and decision type drive this difficulty.

\subsection{Long-Horizon Case Difficulty}
\label{sec:long-horizon}
\begin{figure}
    \centering
    \includegraphics[width=\linewidth]{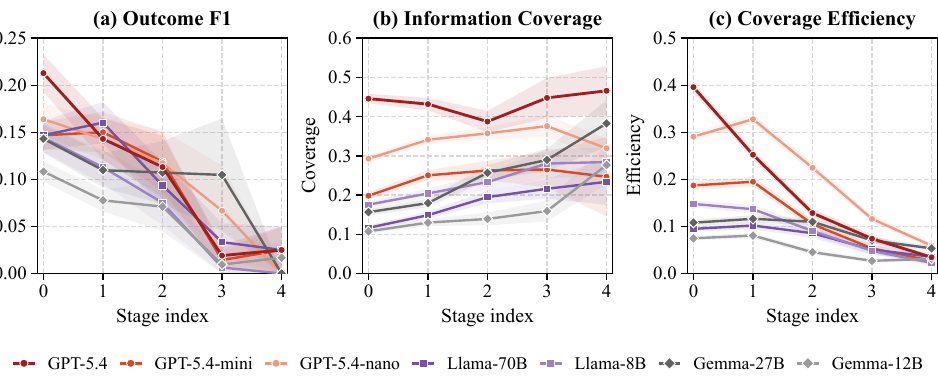}
    \caption{\textbf{Per-stage performance on \ours{}.} Decision F1 (a) declines monotonically with stage index. Information coverage (b) rises; coverage efficiency (c) falls. Shaded bands are standard errors.}
    \label{fig:stage-progression}
    \vspace{-10pt}
\end{figure}

Figure~\ref{fig:stage-progression} traces performance by stage index over management stages, those carrying a medication or procedure decision; we set aside the diagnosis stage that closes every case, whose higher scores (Table~\ref{tab:main-results-long-horizon}) would otherwise mask the management trajectory. Decision F1 falls sharply with stage index for every model (Figure~\ref{fig:stage-progression}a): GPT-5.4 drops from 0.21 at the first management stage to below 0.03 by the fourth, and the others converge toward zero. Later decisions must be reconciled with a longer record of prior actions and observations, and every model finds this progressively harder.

Information access is not the bottleneck. Coverage holds roughly constant for GPT-5.4 and even rises for weaker models as the record grows (Figure~\ref{fig:stage-progression}b), yet decisions still degrade. What collapses instead is efficiency (Figure~\ref{fig:stage-progression}c): GPT-5.4's query count climbs from about four to over fifty per stage while coverage stays near 0.45, so late stages are dominated by redundant queries that surface no new information and yield worse decisions. The binding constraint is therefore late-stage clinical reasoning, not information access, a dissociation that outcome-only evaluation cannot detect.
\subsection{Information Seeking Reduces Waste}
\label{sec:waste}
\begin{figure}
    \centering
    \includegraphics[width=\linewidth]{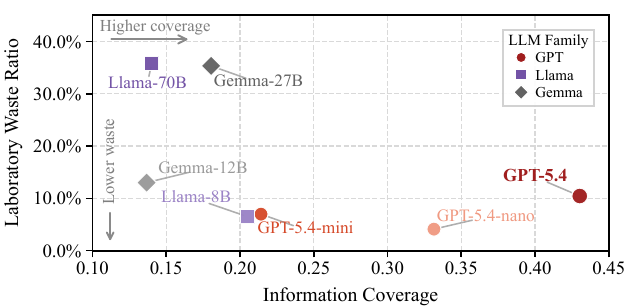}
    \caption{\textbf{Coverage-waste on \ours{}.} Information coverage is inversely related to laboratory waste: models that gather more relevant information also waste less.}
    \label{fig:frontier}
    \vspace{-10pt}
\end{figure}

If gathering information simply meant querying more, higher coverage would buy more wasted tests. Figure~\ref{fig:frontier} shows the reverse: coverage and laboratory waste are inversely related, so models that retrieve more of the relevant information also waste less. GPT-5.4 sits in the favorable corner, pairing the highest coverage with low waste, while Llama-70B and Gemma-27B fall into the opposite region with low coverage and waste above $35\%$. GPT-5.4-nano reaches moderate coverage at the lowest waste of any model ($\sim$4\%), so efficient information seeking does not require scale.

Coverage and waste thus reflect one underlying skill, knowing which information is worth acquiring, rather than two independent axes: a model that knows what to look for both retrieves the relevant evidence and refrains from tests the clinical picture does not warrant. \ours{} measures this skill directly, and the frontier gives a concrete target for clinical agents, the high-coverage, low-waste corner, rather than decision accuracy in isolation.

\subsection{Diagnosis Recovered; Management Not}
\label{sec:type}

The stage analysis points to decision type as the underlying axis of difficulty, which Table~\ref{tab:type-breakdown} makes explicit. On diagnosis-only stages, models recover the discharge diagnoses reasonably well (0.40 pooled, 0.51 for GPT-5.4), but on management stages F1 collapses to 0.13. The gap is roughly threefold and holds for every model: recognizing what the patient has is far easier than committing to what to do about it. This also explains why the appended diagnosis stage inflates whole-case scores and why, once it is removed, the management trajectory in Section~\ref{sec:long-horizon} falls so steeply.

Medications are the limiting case. Every model assigns the correct action type (\texttt{start}, \texttt{stop}, \texttt{switch}, \texttt{adjust}) for 53--73\% of medication decisions, yet medication match scores stay around 0.10 at best. With the action gate largely satisfied, the residual error is one of drug identity: models propose a plausible class of therapy but rarely the specific agent the treating team selected, and ATC partial credit reflects that distance. Action accuracy does not track the final score, Llama-8B has the highest action accuracy (0.73) but one of the lowest medication scores (0.082), confirming that choosing the right drug, not the right action, is the binding difficulty.

This difficulty is not relieved by gathering more information. At the stage level, coverage and decision F1 are essentially uncorrelated (Pearson $r{=}0.08$ overall, $0.04$ within GPT-5.4), so retrieving more of the record does not by itself yield better decisions. What models retrieve is also lopsided: every model consults the nurse channel most and the laboratory least (GPT-5.4 covers 0.89 of nurse facts but 0.25 of lab facts), and the weaker models waste tests because their laboratory queries miss, 60--73\% of their lab orders name tests not present in the stage, against 11--19\% for the GPT models. Competence on \ours{} is thus bounded by clinical judgment, selecting the right management action and the right test, rather than by the ability to retrieve information.

\begin{table}[t]
\centering
\footnotesize
\setlength{\tabcolsep}{4.8pt}
\renewcommand{\arraystretch}{0.9}
\caption{\textbf{Decision-type decomposition.} Diagnosis-only stages score far above management stages. \emph{Lab miss} is the fraction of laboratory queries absent from the stage.}
\label{tab:type-breakdown}
\begin{tabular}{lccccc}
\toprule
\rowcolor{clinblue}
\clinhead{Model} & \clinhead{Diag} & \clinhead{Mgmt} & \clinhead{Action} & \clinhead{Med} & \clinhead{Lab} \\
\rowcolor{clinblue}
& \clinhead{F1} & \clinhead{F1} & \clinhead{acc.} & \clinhead{score} & \clinhead{miss} \\
\midrule
GPT-5.4        & \clinmetric{0.511} & \clinmetric{0.172} & 0.56 & 0.097 & 14\% \\
\rowcolor{clinlight}
GPT-5.4-mini   & 0.440 & 0.139 & 0.57 & 0.083 & 19\% \\
GPT-5.4-nano   & 0.416 & 0.147 & 0.53 & 0.085 & \clinmetric{11\%} \\
\rowcolor{clinlight}
Llama-70B      & 0.358 & 0.140 & 0.70 & \clinmetric{0.101} & 61\% \\
Llama-8B       & 0.325 & 0.121 & \clinmetric{0.73} & 0.082 & 60\% \\
\rowcolor{clinlight}
Gemma-27B      & 0.383 & 0.126 & 0.56 & 0.087 & 73\% \\
Gemma-12B      & 0.360 & 0.090 & 0.68 & 0.070 & 66\% \\
\bottomrule
\end{tabular}
\end{table}

\section{Conclusion}
\label{sec:conclusion}

We presented \ours{}, which evaluates LLMs as attending physicians over real, multi-stage admissions. Under its Longitudinal Inpatient Simulation paradigm, cases are built automatically from EHR timelines, clinical information is withheld until the model actively queries one of four agents, and every decision is scored deterministically alongside process metrics for coverage and cost. Across the evaluated models, the strongest attains only modest decision F1, and outcome quality is decoupled from how efficiently models gather information. Difficulty concentrates in management decisions and later stages, where accuracy collapses even as coverage holds, indicating that clinical reasoning, not information access, is the binding constraint. \ours{} makes this gap, invisible to outcome-only evaluation, directly measurable.

\clearpage
\section{Limitations}

Because \ours{} grounds every decision in the management actually recorded for each admission, it measures concordance with real clinical practice rather than against a single optimal answer; a clinically reasonable alternative may therefore score as a miss. This property is intrinsic to any benchmark derived from real EHR trajectories and bounds how scores should be read, but does not affect the relative comparison of models on the same fixed cases.

Cases come from one US academic medical center and use English notes with US coding and pricing systems (ICD, ATC, the CMS Clinical Laboratory Fee Schedule, and NADAC); generalization to other settings, languages, and coding standards is left to future work. Finally, construction and some scoring steps use LLMs, but all structured ground truth (drug names, ICD codes) is derived deterministically from EHRs rather than from model output, which keeps the evaluation targets independent of the models under test.

\section{Ethical Considerations}

\ours{} is derived from MIMIC-IV and MIMIC-IV-Note, which are de-identified and distributed through PhysioNet under a credentialed data use agreement; we comply with those terms, expose no protected health information, and require users to hold the appropriate PhysioNet credentials. \ours{} is a research benchmark for measuring LLM behavior, not a clinical decision-support tool: the models evaluated here fall far short of safe clinical use, and strong benchmark scores do not imply safe or equitable behavior in practice, particularly given the single-center source. We intend the benchmark to expose capability gaps and thereby discourage premature claims of clinical readiness.

\bibliography{custom}

% Replace placeholder entries in custom.bib with complete, verified references.
\clearpage
\appendix

\section{Extended Related Work}
\label{app:related-extended}

This appendix provides per-benchmark detail and an extended comparison
referenced in Section~\ref{sec:related}. We describe task setup, data
source, and scale for each cited work; the categorical argument and
condensed comparison appear in the main text.

\paragraph{Knowledge benchmarks.}
MedQA~\citep{jin2021disease} contains 12{,}723 USMLE-style
multiple-choice questions; MedMCQA~\citep{pal2022medmcqa} provides
194k questions from Indian medical entrance exams;
PubMedQA~\citep{jin2019pubmedqa} contains 273k yes / no / maybe
questions derived from PubMed abstracts;
MMLU-Health~\citep{hendrycks2020measuring} covers approximately 2k items
across clinical knowledge, anatomy, college medicine, medical
genetics, and professional medicine;
MultiMedQA~\citep{singhal2023large} aggregates these and adds
consumer-facing health QA sets. All are static multiple-choice or
short-answer formats with fully specified vignettes.

\paragraph{EHR query benchmarks.}
EHRSQL~\citep{lee2022ehrsql} provides 24k natural-language questions
paired with SQL queries over MIMIC-III and eICU;
MIMIC-SQL~\citep{wang2020text} contains 10k similar pairs over
MIMIC-III; FHIR-AgentBench~\citep{lee2025fhir} provides 2{,}931
questions over MIMIC-IV-FHIR with both SQL and FHIR-API answers.
Tasks are single-shot translations from natural language to structured
queries, scored by execution match against the EHR database.

\paragraph{MedAgentBench.}
MedAgentBench~\citep{jiang2025medagentbench} provides 300
physician-authored tasks across 10 categories (chart review, order
placement, result retrieval, among others) operating on 100 patient
profiles drawn from Stanford STARR. The environment exposes a
FHIR-compliant API matching modern EMR systems, and success is scored
by post-action database state. Tasks are atomic and pre-specified
rather than emerging from longitudinal patient management.

\paragraph{AgentClinic.}
AgentClinic~\citep{schmidgall2024agentclinic} composes a doctor agent
with an LLM-played patient agent, a measurement agent that returns
test results, and optionally a moderator. Cases are derived from
MedQA and NEJM Image Challenges; the doctor agent must converge on a
single diagnosis through bounded dialogue turns. Evaluation covers
diagnostic accuracy and patient-centric metrics such as compliance
and confidence; intermediate dialogue quality relies on LLM-as-judge.

\paragraph{Other simulated-patient benchmarks.}
MedDialBench~\citep{luo2026meddialbench} introduces 24 parametric adversarial
patient behaviors layered onto consultation cases;
MAQuE~\citep{gong2025dialogue} adds incremental behavioral layers across 3{,}000
simulated patients, measuring each layer's marginal effect on
diagnostic accuracy; HELPMed~\citep{bean2025clinical} evaluates LLMs with real
human participants playing patient roles, reporting dramatic accuracy
drops relative to standalone settings; AMIE~\citep{tu2024towards} is a
diagnostic dialogue system evaluated in randomized OSCE-style
consultations against primary-care physicians.

\paragraph{Coding-style executable benchmarks.}
MedCalc-Env~\citep{mao2025medcalc} is an RL environment built on
the InternBootcamp framework for multi-step medical calculation,
covering 700+ tasks across specialties.
MedAgentGym~\citep{xu2025medagentgym} provides 72k task instances
across 129 categories of biomedical coding scenarios (SQL queries
over MIMIC-III and eICU, calculator-style reasoning, data-analysis
code generation), positioned as a training-time analog of SWE-Gym.
Both target code or formula production as the unit of evaluation
rather than the clinical decisions themselves.

\paragraph{Coding agent benchmarks referenced for analogy.}
HumanEval~\citep{chen2021evaluating} introduced executable
function-level code evaluation via unit tests;
SWE-bench~\citep{jimenez2024swe} scaled this to repository-level
GitHub issue resolution with executable test suites. The
deterministic-verification spirit of these benchmarks informs the
scoring design of \ours{}, though the unit of evaluation differs
(clinical decisions rather than code).

Table~\ref{tab:related-comparison-extended} expands the main-text
comparison with automatic construction and structured-action dimensions.

\begin{table*}[t]
\centering
\footnotesize
\setlength{\tabcolsep}{2.8pt}
\renewcommand{\arraystretch}{1.02}
\caption{\textbf{Extended design-dimension comparison across representative medical LLM benchmarks.}
\cmark{} = supported, $\triangle$ = partial or limited, \xmark{} = not supported.
The seven dimensions extend Table~\ref{tab:related-comparison} with
automatic case construction from raw records and structured action space matched to attending-physician decisions.}
\label{tab:related-comparison-extended}
\begin{tabular}{lccccccc}
\toprule
\rowcolor{clinblue}
\clinhead{Benchmark}
& \clinhead{Real EHR}
& \clinhead{Multi-}
& \clinhead{Auto}
& \clinhead{Active}
& \clinhead{Structured}
& \clinhead{Deterministic}
& \clinhead{Process /} \\
\rowcolor{clinblue}
& \clinhead{trajectory}
& \clinhead{stage}
& \clinhead{construction}
& \clinhead{info-seek}
& \clinhead{action}
& \clinhead{scoring}
& \clinhead{cost} \\
\midrule

\rowcolor{clinlight}
\multicolumn{8}{l}{\textbf{Static medical knowledge benchmarks}} \\
MedQA~\citep{jin2021disease}
& \xmark & \xmark & \xmark & \xmark & \xmark & \cmark & \xmark \\
MedMCQA~\citep{pal2022medmcqa}
& \xmark & \xmark & \xmark & \xmark & \xmark & \cmark & \xmark \\
PubMedQA~\citep{jin2019pubmedqa}
& \xmark & \xmark & \xmark & \xmark & \xmark & \cmark & \xmark \\
MultiMedQA~\citep{singhal2023large}
& \xmark & \xmark & \xmark & \xmark & \xmark & \cmark & \xmark \\

\addlinespace[1pt]
\rowcolor{clinlight}
\multicolumn{8}{l}{\textbf{EHR query and data-access benchmarks}} \\
EHRSQL~\citep{lee2022ehrsql}
& \cmark & \xmark & \cmark & \xmark & $\triangle$ & \cmark & \xmark \\
MIMIC-SQL~\citep{wang2020text}
& \cmark & \xmark & \cmark & \xmark & $\triangle$ & \cmark & \xmark \\
FHIR-AgentBench~\citep{lee2025fhir}
& \cmark & \xmark & \cmark & \xmark & $\triangle$ & \cmark & \xmark \\
MedAgentBench~\citep{jiang2025medagentbench}
& $\triangle$ & \xmark & \xmark & \cmark & \cmark & \cmark & \xmark \\

\addlinespace[1pt]
\rowcolor{clinlight}
\multicolumn{8}{l}{\textbf{Interactive diagnostic and simulated-patient benchmarks}} \\
AgentClinic~\citep{schmidgall2024agentclinic}
& \xmark & \xmark & \xmark & \cmark & $\triangle$ & \xmark & \xmark \\
MedDialBench~\citep{luo2026meddialbench}
& \xmark & \xmark & $\triangle$ & \cmark & \xmark & \xmark & \xmark \\
MAQuE~\citep{gong2025dialogue}
& \xmark & \xmark & $\triangle$ & \cmark & \xmark & \xmark & \xmark \\
HELPMed~\citep{bean2025clinical}
& \xmark & \xmark & \xmark & \cmark & \xmark & \xmark & \xmark \\
AMIE~\citep{tu2024towards}
& \xmark & \xmark & \xmark & \cmark & \xmark & \xmark & \xmark \\

\addlinespace[1pt]
\rowcolor{clinlight}
\multicolumn{8}{l}{\textbf{Executable and coding-style medical agent benchmarks}} \\
MedCalc-Env~\citep{mao2025medcalc}
& \xmark & $\triangle$ & \xmark & \xmark & $\triangle$ & \cmark & \xmark \\
MedAgentGym~\citep{xu2025medagentgym}
& $\triangle$ & \xmark & $\triangle$ & \cmark & $\triangle$ & \cmark & \xmark \\

\midrule
\rowcolor{clinlight}
\textbf{\ours{} (Ours)}
& \clinmetric{\cmark}
& \clinmetric{\cmark}
& \clinmetric{\cmark}
& \clinmetric{\cmark}
& \clinmetric{\cmark}
& \clinmetric{\cmark}
& \clinmetric{\cmark} \\
\bottomrule
\end{tabular}
\end{table*}

\section{Data Preprocessing Details}
\label{app:data-preprocessing}

\subsection{Input Sources}
\label{app:input-sources}

Table~\ref{tab:data-sources} summarizes the source groups used to build the admission timelines. All tables are linked at the admission level using \texttt{subject\_id} and \texttt{hadm\_id}. Rows without a valid admission identifier are excluded before timeline construction. 

\begin{table*}[t]
\centering
\footnotesize
\setlength{\tabcolsep}{7pt}
\renewcommand{\arraystretch}{1.05}
\caption{\textbf{Source groups used in admission-level preprocessing.}
Each row lists a structured or note source, the fields extracted from
it, and how those fields enter the admission-level event timeline that
downstream case construction and agent readviews draw on.}
\label{tab:data-sources}
\begin{tabular}{p{0.16\textwidth}p{0.36\textwidth}p{0.38\textwidth}}
\toprule
\rowcolor{clinblue}
\clinhead{Group} & \clinhead{Tables} & \clinhead{Main retained information} \\
\midrule

Encounter context &
\texttt{hosp\_admissions\_df} &
Admission/discharge timing, admission type, demographics, insurance,
language, and discharge disposition \\

\rowcolor{clinlight}
Laboratory and microbiology &
\texttt{hosp\_labevents\_df}, \texttt{hosp\_microbiologyevents\_df} &
Test labels, values, units, reference ranges, abnormality flags, specimens,
organisms, and comments \\

Medication orders and administrations &
\texttt{hosp\_prescriptions\_df}, \texttt{hosp\_pharmacy\_df},
\texttt{hosp\_emar\_detail\_df} &
Drug names, dose, route, frequency, order status, administration type, and
infusion details \\

\rowcolor{clinlight}
ICU observations and flowsheets &
\texttt{ehr\_chartevents\_df}, \texttt{ehr\_datetime\_events\_df},
\texttt{ehr\_inputevents\_df}, \texttt{ehr\_outputevents\_df},
\texttt{ehr\_ingredientevents\_df}, \texttt{ehr\_procedureevents\_df} &
Vitals, bedside measurements, fluid inputs and outputs, procedure events,
ingredients, rates, and durations \\

Coded outcomes &
\texttt{hosp\_diagnoses\_icd\_df}, \texttt{hosp\_procedures\_icd\_df} &
ICD diagnosis and procedure codes, code versions, and mapped long-title
descriptions \\

\rowcolor{clinlight}
Clinical notes &
\texttt{note\_df}, \texttt{radiology\_note} &
Discharge summaries, narrative hospital course, chief complaint, history,
and radiology report text \\

\bottomrule
\end{tabular}
\end{table*}

\subsection{Timeline Construction}
\label{app:admission-assembly}

For each admission, preprocessing follows deterministic steps that do not use LLMs:
\begin{enumerate}[leftmargin=*,nosep]
  \item discover the selected structured and note CSV files and stream them in chunks;
  \item normalize \texttt{subject\_id} and \texttt{hadm\_id}, then drop rows missing either key;
  \item assign a canonical \texttt{event\_time} using the first available timestamp in this priority order: \texttt{charttime}, \texttt{starttime}, \texttt{admittime}, \texttt{chartdate}, \texttt{stoptime}, \texttt{endtime};
  \item serialize each row as an event with \texttt{source\_table}, \texttt{event\_time} when available, and table-specific payload fields, while removing identifiers and internal processing columns;
  \item group events by admission and sort them by timestamp, \texttt{source\_table}, and original row order for deterministic tie breaking.
\end{enumerate}

The preprocessor writes one JSON timeline per admission under subject/admission-specific directories. Each file is a JSON array of ordered events. In the preprocessing release used by this work, the resulting timeline collection contains 229,482 admission files across 24,760 subjects and 237,814,500 retained events.

\subsection{Event Schema Example}
\label{app:event-schema}

Each event keeps provenance through \texttt{source\_table} and preserves only fields meaningful for the source row. Table~\ref{tab:event-schema} shows schematic examples; patient-specific note text is omitted.

\begin{table*}[t]
\centering
\footnotesize
\setlength{\tabcolsep}{7pt}
\renewcommand{\arraystretch}{1.05}
\caption{\textbf{Timeline event schemas after preprocessing.}
Representative events from each source. Every event preserves its source
provenance and the source-specific clinical fields later used for case
construction, agent readviews, and ground-truth scoring.}
\label{tab:event-schema}
\begin{tabular}{p{0.18\textwidth}p{0.34\textwidth}p{0.38\textwidth}}
\toprule
\rowcolor{clinblue}
\clinhead{Event type} & \clinhead{Example retained fields} & \clinhead{Use in \ours{}} \\
\midrule

Admission &
\texttt{source\_table}, \texttt{event\_time}, admission type, race, gender, age &
Initial patient context and admission boundary \\

\rowcolor{clinlight}
Lab result &
\texttt{source\_table}, \texttt{event\_time}, label, value, unit,
reference range, flag &
Lab-agent responses and evidence for clinical stages \\

Medication &
\texttt{source\_table}, \texttt{event\_time}, drug or medication name,
dose, route, frequency, status &
Medication ground-truth enrichment and nurse/pharmacy evidence \\

\rowcolor{clinlight}
Note &
\texttt{source\_table}, \texttt{event\_time}, text &
Discharge-note decision extraction and prior-admission history \\

ICD code &
\texttt{source\_table}, ICD code, ICD version, long-title description &
Diagnosis and procedure ground truth \\

\bottomrule
\end{tabular}
\end{table*}

\subsection{Worked Preprocessing Example}
\label{app:worked-preprocessing-example}

Consider an admission with an admission row at $t_0$, laboratory results at $t_1$, a medication order at $t_2$, a discharge note at $t_3$, and ICD-coded diagnoses assigned to the same \texttt{hadm\_id}. The preprocessor converts these heterogeneous rows into the ordered sequence: 
\begin{equation}
\scriptsize
  [e_{\mathrm{adm}}(t_0),\; e_{\mathrm{lab}}(t_1),\;
  e_{\mathrm{med}}(t_2),\; e_{\mathrm{note}}(t_3),\;
  e_{\mathrm{icd}}]
\end{equation}
where each $e$ retains its original \texttt{source\_table}. Later pipeline stages use the same sequence differently: Phase A reads the discharge-note event to extract decisions; Phase B anchors those decisions against the ordered timeline; Phase C partitions timeline intervals into stage contexts; and Phase D enriches ground truth from the structured medication, diagnosis, and procedure events.

This separation keeps data preprocessing independent from benchmark scoring. The timeline builder standardizes timing and provenance, while the case-construction pipeline decides which events become patient, nurse, lab, and history readviews and which structured records define evaluation targets.

\section{Source-Level Implementation Details}
\label{app:source-implementation}

This appendix records implementation details from the released \ours{}
source pipeline. The goal is to make the benchmark construction and
evaluation process auditable without moving code-level detail into the main
paper.

\subsection{Case Generation Artifacts}
\label{app:case-artifacts}

The pipeline persists enough state to audit each case without inspecting
the raw database again. The planning record stores extracted decisions,
timeline anchors, stage context bounds, and enrichment outputs. The
prepared case record stores the executable stage sequence, role-specific
readviews, available agents, submission hints, and held-out
ground-truth items. The model-execution record stores the full dialogue
and submitted decisions for each stage, and the evaluation record stores
matching assignments and process metrics. The case manifest contains one
JSONL record per prepared admission with \texttt{subject\_id},
\texttt{hadm\_id}, and the number of constructed stages.
Table~\ref{tab:source-artifacts} summarizes these retained artifacts.

\begin{table*}[t]
\centering
\footnotesize
\setlength{\tabcolsep}{7pt}
\renewcommand{\arraystretch}{1.04}
\caption{\textbf{Auditable artifacts retained by the \ours{} construction
pipeline.} Each layer persists the reader-visible information needed to
reconstruct, inspect, or score a benchmark item, making every case fully
traceable back to its source EHR.}
\label{tab:source-artifacts}
\begin{tabular}{p{0.18\textwidth}p{0.74\textwidth}}
\toprule
\rowcolor{clinblue}
\clinhead{Layer}& \clinhead{Retained audit information} \\
\midrule

Admission loading &
Separates admission metadata, discharge note, ICD diagnosis records, and
timestamped clinical events; assigns sequential event indices. \\

\rowcolor{clinlight}
Prior history &
Collects same-patient admissions before the current hospitalization and
stores both summary fields and full discharge-note detail. \\

Planning &
Extracts decisions, anchors them to timeline events, constructs stages,
merges very small contexts, and enriches ground truth from structured
records. \\

\rowcolor{clinlight}
Readviews &
Builds cumulative patient, nurse, lab, and history views for each stage
while preserving stage boundaries. \\

Runtime &
Runs direct or interactive episodes, exposes only active tools, records
submissions and dialogue traces. \\

\rowcolor{clinlight}
Evaluation &
Computes ontology-grounded outcome scores and process metrics for
information coverage, laboratory cost, and medication cost. \\

\bottomrule
\end{tabular}
\end{table*}

\subsection{Admission Loading and Timeline Separation}
\label{app:source-loading}

The source loader intentionally separates static admission-level records
from the clinical event timeline. Admission rows become metadata;
diagnosis-code rows become held-out diagnosis candidates; discharge-note
rows become narrative context for decision extraction. All other source
rows enter the event timeline with a sequential \texttt{index},
\texttt{event\_time}, \texttt{source\_table}, and source-specific
payload. Events before admission time are retained but flagged as
\texttt{pre\_admission}; anchoring later rejects decisions that can only
be grounded before the admission begins. This separation prevents coded
outcomes and discharge summaries from being passively exposed to the
model while still making them available for planning and scoring.

Prior admissions are loaded only for the same patient and only when their
admission time precedes the current admission. The history view stores a
summary level: admission and discharge time, chief complaint, brief
hospital course, and discharge diagnosis, plus a full-note detail level
retrievable by index through the history tool.

\subsection{Planner Safeguards}
\label{app:planner-safeguards}

The planner includes several implementation safeguards that are easy to
miss from the main description:

\begin{itemize}[leftmargin=*,nosep]
  \item \textbf{Sequential anchoring:} Phase B uses a forward search over
        20-event windows. The next decision begins after the previous
        accepted anchor, preserving the temporal order extracted from the
        discharge note.
  \item \textbf{Leakage expansion:} after a single anchor is found, a
        local scan examines up to 10 events before and after the anchor.
        Neighboring records for the same drug, procedure, or action are
        absorbed into the anchor range so that these records are not
        included as pre-decision evidence.
  \item \textbf{Continuation filtering:} medication decisions are skipped
        when the same drug already appears as active medication in the
        proposed context window, preventing routine continuation orders
        from becoming artificial decisions.
  \item \textbf{Small-context merging:} stages with fewer than 10 context
        events are merged into an adjacent stage when possible, reducing
        nearly empty decision points that would be hard to evaluate
        meaningfully.
  \item \textbf{Groundable diagnoses:} diagnosis ground truth is filtered
        through a separate sliding-window scan over the event timeline
        using 100-event windows and 50-event stride; only ICD codes judged
        evidenced or managed in at least one window are retained.
\end{itemize}

After stages are constructed, Phase D converts raw decisions to
structured \texttt{gts}. Medication fields such as drug name, dose, route,
frequency, product strength, and order status are copied from
prescription, pharmacy, or eMAR records. Procedure and diagnosis fields
are copied from ICD-coded rows, including code, version, and long-title
description. LLM calls classify decision type and medication action, but
the final evaluable fields are copied from structured source records.

\subsection{Information Agent Specifications}
\label{app:agent-details}

Each information agent is powered by GPT-5.4-mini and restricted to a
designated readview $\mathcal{V}_a$, a filtered projection of the EHR events
relevant to that role.

\paragraph{Patient Agent.}
The patient agent simulates subjective reporting from the patient's
perspective. Given a query, it draws from demographics, chief complaint (CC),
history of present illness (HPI), and past medical history (PMH) to produce
lay-language responses that reflect how a real patient might describe their
symptoms and background, without access to objective clinical measurements.

\paragraph{Nurse Agent.}
The nurse agent reports bedside observations from the nursing record. It has
access to vital signs, fluid balance, and medication administration records
(MAR), and responds to queries about the patient's current physiological
status and treatment as observed at the bedside.

\paragraph{Lab Agent.}
The lab agent does not expose results passively. Given a natural language query
$q$ and the set of available test labels $\mathcal{L}_k$ in stage
$\mathcal{S}_k$, it identifies the matching subset $\mathcal{L}' \subseteq
\mathcal{L}_k$ via LLM-based semantic matching that handles abbreviations,
panel expansions (e.g., CBC, BMP), and synonyms, then returns value, unit, and
reference range for each matched item. Tests not in $\mathcal{L}_k$ return a
null response.

\paragraph{History Agent.}
The history agent supplies longitudinal context from prior discharge summaries.
It is activated only when the patient has at least one prior admission on
record, preventing uninformative interactions when no historical data exists.

\paragraph{Agent Activation.}
Agent $a$ is activated in stage $\mathcal{S}_k$ only if its readview $\mathcal{V}_a$ intersects the current context window, preventing uninformative interactions with agents that have no relevant data for the stage. The history agent is additionally gated on whether the patient has any prior admissions.

\subsection{Readview Construction and Tool Exposure}
\label{app:readview-source}

Readviews are cumulative within an admission: at stage $k$, each role can
see records from event index 0 through the stage end, filtered to that
role's allowed source tables. The patient readview contains static
demographics, chief complaint, history of present illness, and past
medical history. The nurse readview contains bedside observations,
charted events, fluid input and output, administered medications,
pharmacy and prescription records, radiology summaries, and procedure
events. The lab readview contains laboratory and microbiology events,
excluding degenerate test labels shorter than two characters.

Tool exposure is stage-specific. The patient and history roles are
available by default when their data exist, while nurse and lab tools are
activated only when the current stage window contains matching events.
During interactive evaluation, only submit tools corresponding to
ground-truth types in the current stage are exposed.
Duplicate submissions with the same type, value, and medication action
are ignored by the runtime.

\subsection{Runtime State Across Stages}
\label{app:runtime-state}

The interactive runtime allows at most 60 turns per stage and permits one
tool call per turn. To keep the longitudinal state clinically coherent,
each new stage receives filtered prior-stage context: information-seeking
dialogue is retained, while prior submit and finalize calls are removed.
The environment then injects the previous stage's confirmed ground-truth
decisions into the context, so later stages condition on what actually
happened in the hospitalization rather than on earlier model errors. In
direct mode, the same information is shown as prior realized events.

\subsection{Evaluation Implementation Details}
\label{app:source-evaluation}

Outcome evaluation uses type-wise Hungarian matching. Diagnosis and
procedure submissions are mapped to ICD codes by embedding retrieval over
ICD-9-CM, ICD-10-CM, ICD-9-PCS, and ICD-10-PCS vocabularies followed by
LLM reranking, then scored with hierarchical F1 over code ancestors.
Medication submissions are action-gated before drug similarity is
computed: mismatched actions score zero, while action-consistent pairs
receive ATC-hierarchy partial credit. For \texttt{adjust} decisions,
opposing explicit directions halve the drug score.

Process evaluation is computed from the saved dialogue trace. Information
coverage is judged separately for patient, nurse, and lab responses by
comparing retrieved dialogue content against each role's complete
readview, then macro-averaged across roles. The implemented efficiency
metric is the coverage value multiplied by $N/(N+K)$, where $N$ is the
number of judged key facts and $K$ is the number of information-gathering
tool calls. Laboratory cost scoring parses lab tool outputs, counts
unresolved or unavailable lab requests as wasted queries, and estimates
test cost by matching ordered tests to the CMS Clinical Laboratory Fee
Schedule. Medication-cost scoring resolves submitted drugs through
RxNorm, maps to NDC unit prices from NADAC, obtains defined daily doses
from the WHO ATC/DDD index when possible, and reports estimated daily
drug acquisition cost for submitted medication decisions.

\begin{figure*}[t]
\centering
\includegraphics[width=\textwidth]{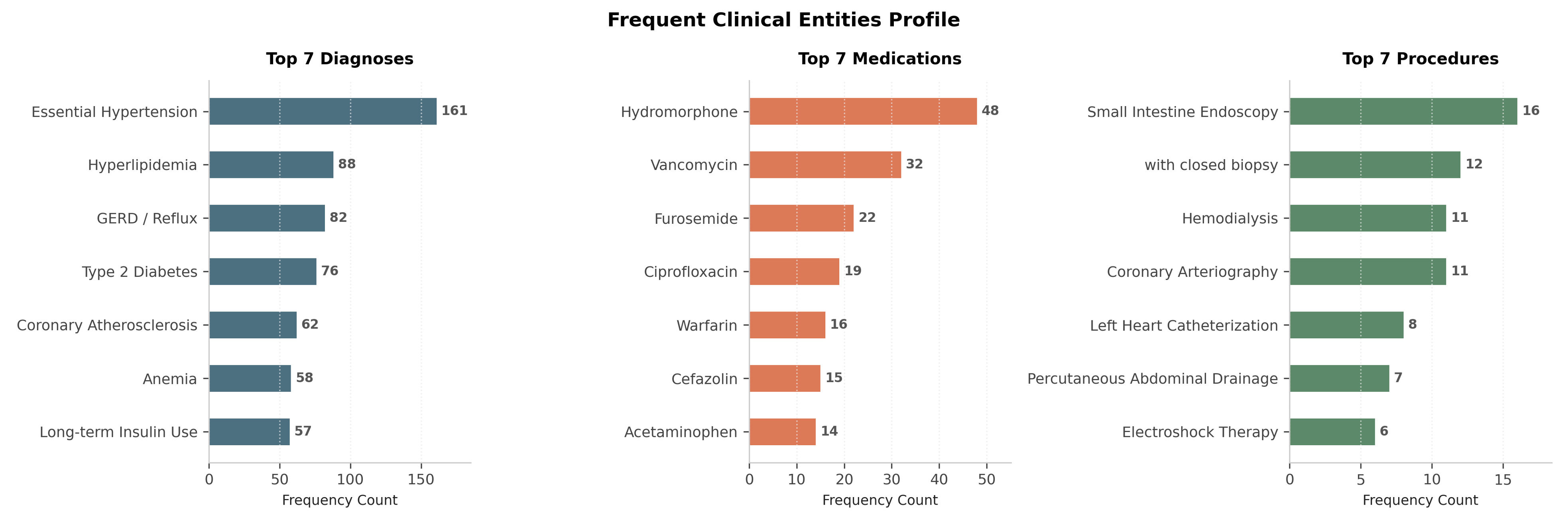}
\caption{\textbf{Frequent clinical entities in \ours{}.} Top diagnosis,
medication, and procedure entities among structured ground-truth
decisions.}
\label{fig:entity-profile}
\end{figure*}

\section{\ours{} Statistics}
\label{app:benchmark-stats}

This appendix details the sampling protocol and construction-quality checks for \ours{}; its composition is summarized in Table~\ref{tab:benchmark-stats} (Section~\ref{sec:benchmark}).

\subsection{Sampling Protocol}
\label{app:sampling}

\ours{} contains 3{,}509 cases, one per admission, drawn from 1{,}809 patients. We discard admissions that yield no valid decision stage after construction, then sample the remainder to span the full range of case horizons, from single-stage admissions to trajectories of up to nine stages. Admissions with no usable discharge note, or with no groundable decisions after Phase~C enrichment, are excluded.

\subsection{Construction Quality}
\label{app:quality}

\ours{} is constructed without manual annotation, so the integrity of
each stage rests on three properties that are guaranteed by construction
and remain verifiable against the source record:

\begin{itemize}[leftmargin=*,nosep]
  \item \textbf{Decision precision:} every extracted decision corresponds
        to an attending-level action actually recorded during the
        admission (a prescription, procedure, or coded discharge
        diagnosis), rather than a hallucinated or non-decision item.
  \item \textbf{Anchor validity:} each decision carries the timeline
        event index it was extracted from, placing it at the point in
        the admission where the action was taken.
  \item \textbf{Enrichment accuracy:} the structured ground-truth fields
        (drug name, action type, dose, ICD code) are copied directly
        from the corresponding table row, not generated.
\end{itemize}

\noindent
Because every stage is built deterministically from logged events, the
ground truth can always be traced back to the originating row.
Appendix~\ref{app:worked-example} presents a complete worked example for
a single admission, showing the information environment the model
observes and the held-out ground truth at each stage.

\subsection{Clinical entities follow common inpatient patterns.}
Figure~\ref{fig:entity-profile} shows the most frequent diagnosis,
medication, and procedure entities in the benchmark. Chronic inpatient
comorbidities such as essential hypertension, hyperlipidemia, GERD, and
diabetes dominate the diagnosis distribution, while medication and
procedure decisions are more dispersed.

\subsection{Longer timelines are less decision-dense.}
\begin{figure*}[t]
\centering
\includegraphics[width=\textwidth]{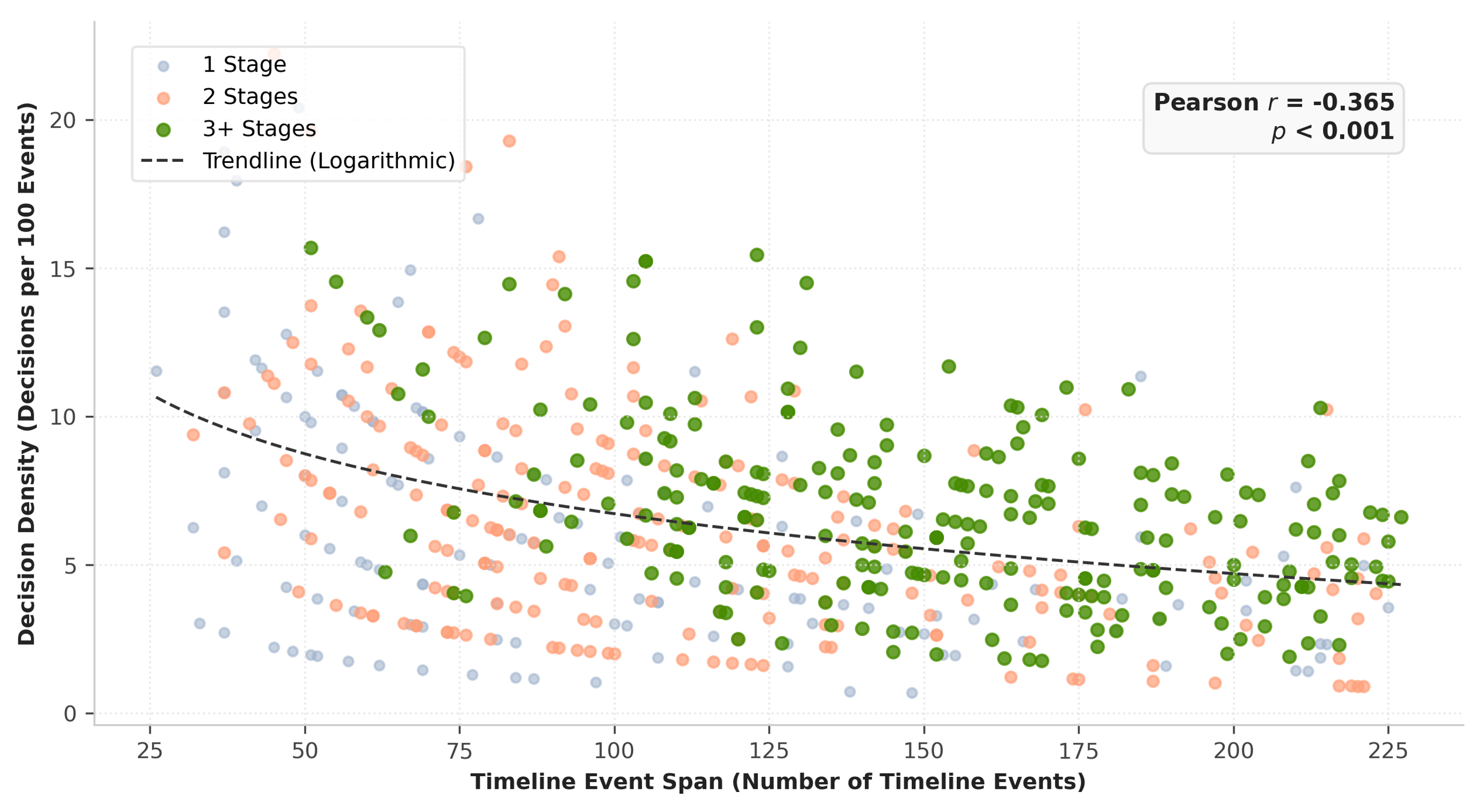}
\caption{\textbf{Decision density by timeline span.} Each point is an
admission, colored by constructed case horizon. Longer event timelines
are associated with lower decision density, so long-horizon cases require
models to reason over more observations per clinical decision.}
\label{fig:span-density}
\end{figure*}

Figure~\ref{fig:span-density} plots each case's timeline event span
against its decision density, colored by case horizon. Decision density
declines as timelines lengthen (Pearson $r=-0.365$, $p<0.001$), showing
that longer admissions add more context per decision rather than simply
more decisions.

\section{\ours{} Example}
\label{app:worked-example}

To make the construction concrete, we trace a single admission end to end.
We use admission \texttt{21569907} from MIMIC-IV, the same chest-pain
and anemia case used in Appendix~\ref{app:dialogue-examples}. The
pipeline segments this admission into three sequential decision stages.
We first show the source admission and the multi-source
information environment exposed to the model, then walk through each
stage: what the model can observe, the decision it must commit, and the
held-out ground truth it is scored against. Patient-facing fields are
drawn from the stored readviews, and structured ground-truth fields are
copied from the paired case record.

\begin{clinbox}{Source admission \texttt{10146904/21569907}}
\footnotesize
\textbf{Demographics.} Female with coronary artery disease, hypertension,
hypercholesterolemia, GERD, prior anemia, diverticulosis, and prior EGD
noted in the history.\\[2pt]
\textbf{Chief complaint.} Chest wall pain and fatigue.\\[2pt]
\textbf{HPI (excerpt).} ``\dots a month-long course of chest pain that
has worsened over the last week\dots fatigued and short of breath\dots
slightly lightheaded\dots pain is located under her left breast and is
worse with palpation\dots received 2 units pRBCs because of a Hct of
22.9\dots denies bleeding, including hematochezia, melena, hematuria or
vomitus streaked with blood\dots''\\[2pt]
\textbf{History agent.} The runtime may expose prior-admission context
through the history tool, but this trace has no usable prior-admission
summary detail.\\[2pt]
\textbf{Held-out decisions.} The admission contains three held-out
decisions: EGD, oral iron supplementation, and later tramadol for rib
pain control.
\end{clinbox}

\begin{clinbox}{Information environment at Stage 0}
\footnotesize
The model does not receive the raw timeline. Instead, at each stage it
may query four specialized agents, each returning a structured readview:
\begin{itemize}[leftmargin=1.2em,nosep,topsep=2pt]
  \item \textbf{Patient:} chief complaint, HPI, past medical history,
        and sex.
  \item \textbf{Nurse:} active medication administrations and nursing
        events, including packed-cell transfusion coding, chest imaging,
        pantoprazole, tramadol, acetaminophen, home cardiac medications,
        and bowel-preparation orders.
  \item \textbf{Lab:} cardiac markers and serial CBC results, including
        hemoglobin 9.2--9.9\,g/dL, MCV 76\,fL, and elevated RDW during
        Stage~0.
  \item \textbf{History:} prior-admission lookup, which may return no
        usable summary for this patient.
\end{itemize}
Coverage and laboratory-cost metrics (Section~\ref{sec:evaluation}) are
computed over exactly which of these facts the model chooses to retrieve
before committing each decision.
\end{clinbox}

\begin{clinstage}{Stage 0 \quad Context events [0,\,58] $\rightarrow$ anchor event 59}
\footnotesize
\textbf{What the model sees.} Early chest-pain evaluation and anemia
workup: a packed-cell transfusion code, chest radiograph with no acute
pulmonary process, chest CT with no PE, medication records including
pantoprazole and tramadol, serial microcytic anemia labs, and a
Golytely bowel-preparation order.\\[2pt]
\textbf{Decision to commit.} A procedure decision for evaluation of
suspected occult GI blood loss / iron-deficiency anemia.\\[2pt]
\textbf{Held-out ground truth.} \emph{Procedure}: ``Performed
esophagogastroduodenoscopy (EGD).'' \quad Enrichment (from
\texttt{hosp\_procedures\_icd}): \texttt{ICD-9 4513}, ``Other
endoscopy of small intestine.''
\end{clinstage}

\vspace{4pt}
\begin{clinstage}{Stage 1 \quad Context events [61,\,104] $\rightarrow$ anchor event 106}
\footnotesize
\textbf{What the model sees.} The admission has progressed past the
endoscopic procedure. The Stage~1 window contains follow-up coagulation,
chemistry, and hematology results, including hemoglobin 9.5\,g/dL,
MCV 74\,fL, hypochromia, microcytes, and potassium replacement.\\[2pt]
\textbf{Decision to commit.} A medication decision to treat persistent
microcytic anemia.\\[2pt]
\textbf{Held-out ground truth.} \emph{Medication} / \texttt{start}:
``Started iron supplements.'' \quad Enrichment (from
\texttt{hosp\_prescriptions}): \texttt{Ferrous Gluconate}, 325\,mg,
route PO, 325\,mg tablet.
\end{clinstage}

\vspace{4pt}
\begin{clinstage}{Stage 2 \quad Context events [107,\,145] $\rightarrow$ anchor event 147}
\footnotesize
\textbf{What the model sees.} Later inpatient medication and lab context:
sucralfate orders, another chemistry panel, hemoglobin 9.4\,g/dL,
MCV 74\,fL, coagulation labs, and scheduled acetaminophen.\\[2pt]
\textbf{Decision to commit.} A medication decision for ongoing rib or
chest wall pain control after the initial chest-pain/anemia workup.\\[2pt]
\textbf{Held-out ground truth.} \emph{Medication} / \texttt{start}:
``Started tramadol for rib pain control.'' \quad Enrichment (from
\texttt{hosp\_prescriptions}): \texttt{TraMADOL (Ultram)}, 50\,mg,
route PO, 50\,mg tablet.
\end{clinstage}

\vspace{4pt}
\noindent
This example illustrates the three properties of
Appendix~\ref{app:quality}: each decision is an action the attending
actually took (the EGD, ferrous gluconate order, and tramadol order),
each is anchored to the exact timeline position it occurred at, and each
structured field is copied from the corresponding row. The
clinical thread is also coherent: the same microcytic-anemia signal that
appears in the Stage~0 dialogue motivates the EGD and the later iron
supplement decision, while Stage~2 shifts to symptom control.

\section{Representative Dialogue Traces}
\label{app:dialogue-examples}

\begin{figure*}[t]
\centering
\includegraphics[width=\textwidth]{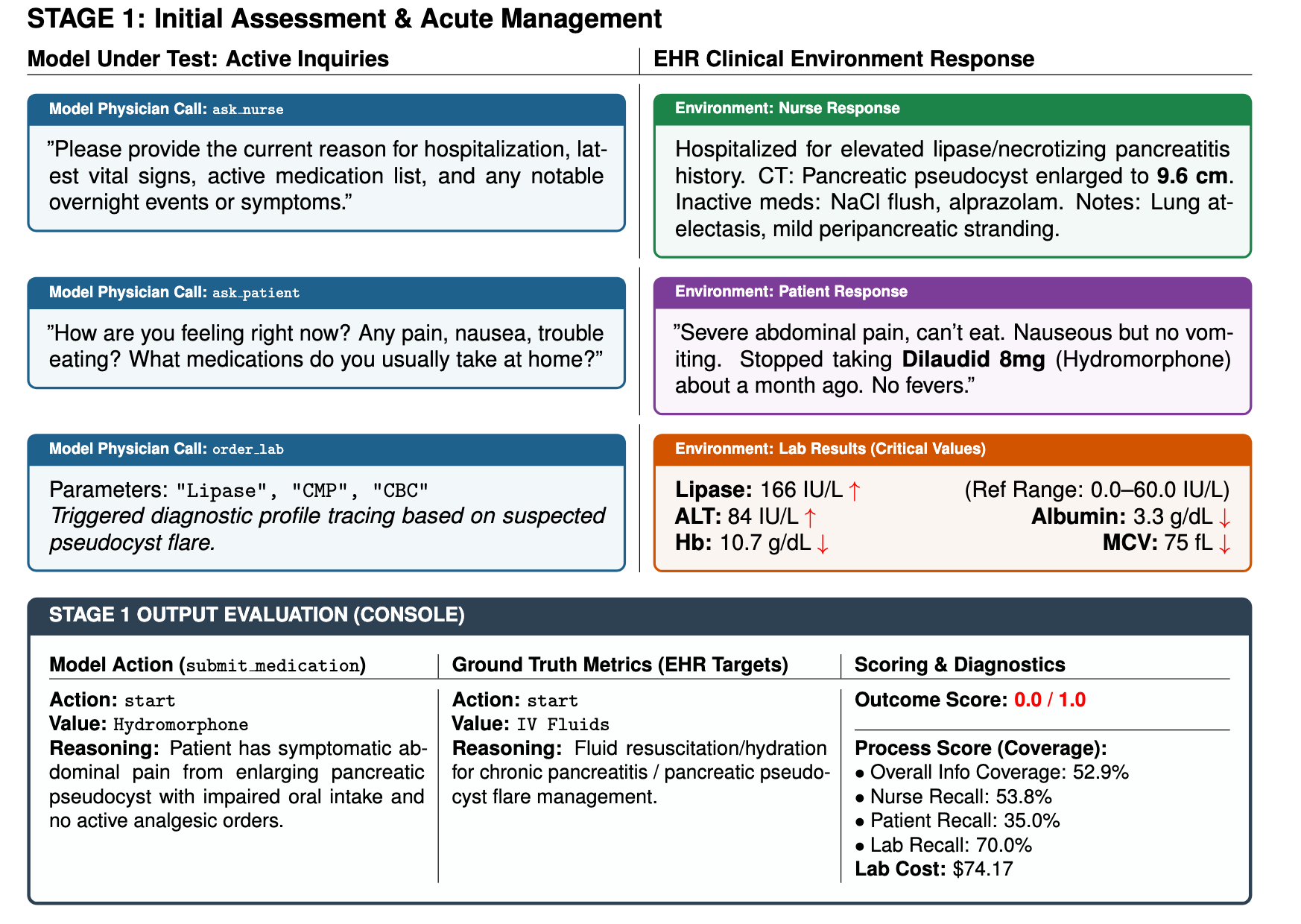}
\caption{\textbf{Runtime stage-level evaluation example.} The model
actively queries nurse, patient, and laboratory agents before submitting
a medication action. The environment records both the outcome mismatch
against the held-out EHR target and process diagnostics such as
information coverage and laboratory cost.}
\label{fig:runtime-stage-example}
\end{figure*}

The worked example above shows the construction-time artifact for a
single admission. Here we show the complementary runtime artifact: a
saved interactive dialogue in which the model must ask role-specific
agents for information before submitting structured decisions.
Figure~\ref{fig:runtime-stage-example} gives a visual example of the
same runtime interface: active model inquiries, environment responses,
submitted action, held-out target, and process diagnostics are all
recorded at the stage level.

\begin{clinbox}{Dialogue trace source \texttt{10146904/21569907}}
\footnotesize
\textbf{Run.} \texttt{gpt5.4}, interactive mode.\\[2pt]
\textbf{Constructed stages.} The paired case record contains three
stages: Stage~0 requires one procedure decision, Stage~1 requires one
medication decision, and Stage~2 requires one medication decision.\\[2pt]
\textbf{Why this trace is useful.} The model queries the patient, nurse,
laboratory, and history tools; the trace includes both successful
retrievals and unavailable requests, making it a compact example of how
\ours{} scores both decisions and information-seeking process.
\end{clinbox}

\begin{clinstage}{Dialogue example A \quad Stage 0 procedure decision}
\footnotesize
\textbf{Required submission.} One procedure.\\[2pt]
\textbf{Transcript excerpt.}\\[-4pt]
\begin{quote}
\footnotesize
\textbf{Model $\rightarrow$ \texttt{ask\_patient}:} ``Hello, I'm
Dr. [Name], one of the attending physicians. Can you tell me what
brought you to the hospital today? What symptoms are you
experiencing?''\\[2pt]
\textbf{Patient $\rightarrow$ Model:} ``I came in because I've been
having chest pain for about a month, and it got worse over the last
week. I've also been feeling really tired and a little short of breath.
This morning I just didn't feel like myself and felt a bit lightheaded,
so I called my doctor and they told me to come in.''\\[2pt]
\textbf{Model $\rightarrow$ \texttt{order\_lab}:} Complete Blood Count
(CBC).\\[2pt]
\textbf{Lab $\rightarrow$ Model:} Hemoglobin 9.2\,g/dL, hematocrit
29.3\%, MCV 76\,fL, RDW 16.1\%; repeat hemoglobin 9.9\,g/dL.\\[2pt]
\textbf{Model $\rightarrow$ \texttt{order\_lab}:} Iron studies.\\[2pt]
\textbf{Lab $\rightarrow$ Model:} ``No results available for 'Iron
studies'. The test was not ordered or did not result for this
patient.''\\[2pt]
\textbf{Model $\rightarrow$ \texttt{ask\_patient}:} ``Have you had any
episodes of nausea, vomiting (especially coffee-ground looking vomit),
or any dark/tarry stools? Also, have you been feeling more tired than
usual or had any shortness of breath with activity?''\\[2pt]
\textbf{Patient $\rightarrow$ model:} ``No, I haven't had any nausea or
vomiting, and I haven't noticed any coffee-ground looking vomit or dark,
tarry stools. I have been feeling more tired than usual, and I've also
had some shortness of breath.''\\[2pt]
\textbf{Model $\rightarrow$ \texttt{submit\_procedure}:}
\texttt{procedure\_name =} ``Esophagogastroduodenoscopy.''\\[2pt]
\textbf{Environment $\rightarrow$ model:} ``status: recorded; type:
procedure; remaining: all done.''
\end{quote}
\textbf{Held-out ground truth.} \emph{Procedure}: ``Performed
esophagogastroduodenoscopy (EGD).''
\end{clinstage}

\vspace{4pt}
\begin{clinstage}{Dialogue example B \quad Stage 1 medication decision}
\footnotesize
\textbf{Required submission.} One medication.\\[2pt]
\textbf{Transcript excerpt.}\\[-4pt]
\begin{quote}
\footnotesize
\textbf{Model $\rightarrow$ \texttt{ask\_patient}:} ``How are you
feeling now? Has the chest pain improved? Have you had any procedures
done since you've been here, like an endoscopy (a camera down your
throat to look at your stomach)?''\\[2pt]
\textbf{Patient $\rightarrow$ model:} ``I'm feeling better now. I'm not
having the chest pain right now, and it has improved since I first came
in. I don't think I've had an endoscopy since I've been here.''\\[2pt]
\textbf{Model $\rightarrow$ \texttt{order\_lab}:} Iron studies (Iron,
TIBC, Ferritin).\\[2pt]
\textbf{Lab $\rightarrow$ Model:} ``No results available for 'Iron
studies (Iron, TIBC, Ferritin)'. The test was not ordered or did not
result for this patient.''\\[2pt]
\textbf{Model $\rightarrow$ \texttt{order\_lab}:} Hemoglobin.\\[2pt]
\textbf{Lab $\rightarrow$ Model:} Hemoglobin 9.2\,g/dL, 9.9\,g/dL, and
9.5\,g/dL on serial measurements.\\[2pt]
\textbf{Model $\rightarrow$ \texttt{ask\_nurse}:} ``Has the patient had
any procedures done during this admission? Specifically, has an EGD
(upper endoscopy) or colonoscopy been performed? Also, has the patient
had any stool guaiac or fecal occult blood test done?''\\[2pt]
\textbf{Nurse $\rightarrow$ Model:} ``Based on the bedside data
available, I do not see documentation of an EGD or colonoscopy being
performed during this admission.''\\[2pt]
\textbf{Model $\rightarrow$ \texttt{submit\_medication}:}
\texttt{action = start, drug\_name =} ``Ferrous Sulfate.''\\[2pt]
\textbf{Environment $\rightarrow$ model:} ``status: recorded; type:
medication; action: start; value: Ferrous Sulfate; remaining: all done.''
\end{quote}
\textbf{Held-out ground truth.} \emph{Medication} / \texttt{start}:
\texttt{Ferrous Gluconate}. This example illustrates why the medication
scorer uses ontology-aware drug matching rather than exact string
identity alone, and why unavailable lab requests contribute to process
cost metrics.
\end{clinstage}

\section{Additional Result Analyses}
\label{app:additional-results}

\paragraph{Stage position explains the apparent late-stage rebound.} Table~\ref{tab:app-stage-position} reports stage-position averages over the benchmark. In multi-stage cases, first and intermediate stages are predominantly management decisions: 98.87\% first-stage rows and 99.07\% intermediate rows contain medication or procedure ground truth. Final stages are predominantly diagnostic: 89.31\% final-stage rows are diagnosis-only. As a result, all-stage averages can rebound at the final stage even though management stages become harder with longer context.

\begin{table}[h]
\centering
\footnotesize
\setlength{\tabcolsep}{7pt}
\renewcommand{\arraystretch}{1.0}
\caption{\textbf{Decision F1 by stage position.} \emph{Single} denotes
single-stage cases; \emph{First}, \emph{Interm.}, and \emph{Final} are positions
within multi-stage cases. Final-stage scores rebound because final stages are
overwhelmingly diagnosis-only ($89.3\%$), whereas first and intermediate stages
are almost entirely management decisions.}
\label{tab:app-stage-position}
\begin{tabular}{lcccc}
\toprule
\rowcolor{clinblue}
\clinhead{Model} & \clinhead{Single} & \clinhead{First} &
\clinhead{Interm.} & \clinhead{Final} \\
\midrule
GPT-5.4        & 0.499 & 0.214 & 0.115 & 0.473 \\
\rowcolor{clinlight}
GPT-5.4-mini   & 0.384 & 0.147 & 0.118 & 0.429 \\
GPT-5.4-nano   & 0.356 & 0.165 & 0.134 & 0.397 \\
\rowcolor{clinlight}
Llama-3.1-70B  & 0.322 & 0.148 & 0.124 & 0.346 \\
Llama-3.1-8B   & 0.297 & 0.145 & 0.094 & 0.305 \\
\rowcolor{clinlight}
Gemma-3-27B    & 0.341 & 0.141 & 0.107 & 0.367 \\
Gemma-3-12B    & 0.334 & 0.109 & 0.073 & 0.330 \\
\bottomrule
\end{tabular}
\end{table}

\paragraph{Long-horizon degradation is consistent across LLM families.} Every model loses decision F1 on admissions with at least three stages (Table~\ref{tab:app-long-degradation}). The absolute drop ranges from 0.029 to 0.072 F1, while interaction length increases for every model. The largest turn increase is for Llama-3.1-70B (+4.0 turns), but this extra interaction does not translate into better coverage or lower waste
in the main results.

\begin{table}[h]
\centering
\footnotesize
\setlength{\tabcolsep}{9pt}
\renewcommand{\arraystretch}{1.0}
\caption{\textbf{Long-horizon degradation relative to all evaluated stages.}
$\Delta F_1$ and relative drop compare decision F1 on admissions with at least
three stages against the full set; $\Delta$ turns is the corresponding change in
interaction length. Every model loses accuracy and takes more turns, and the
extra interaction does not recover the lost performance.}
\label{tab:app-long-degradation}
\begin{tabular}{lccc}
\toprule
\rowcolor{clinblue}
\clinhead{Model} & \clinhead{$\Delta F_1$} &
\clinhead{Rel. drop} & \clinhead{$\Delta$ turns} \\
\midrule
GPT-5.4        & 0.072 & 23.4\% & +2.9 \\
\rowcolor{clinlight}
GPT-5.4-mini   & 0.045 & 17.6\% & +2.1 \\
GPT-5.4-nano   & 0.039 & 15.5\% & +1.4 \\
\rowcolor{clinlight}
Llama-3.1-70B  & 0.033 & 14.4\% & +4.0 \\
Llama-3.1-8B   & 0.029 & 14.6\% & +2.4 \\
\rowcolor{clinlight}
Gemma-3-27B    & 0.033 & 14.5\% & +1.2 \\
Gemma-3-12B    & 0.033 & 16.8\% & +2.0 \\
\bottomrule
\end{tabular}
\end{table}

\paragraph{Decision type remains the dominant accuracy gap.} Averaged across models, diagnosis matching is much stronger than
management matching: mean diagnosis score is 0.412, mean procedure score is 0.284, and mean medication score is only 0.086. Even the best medication score is 0.101 (Llama-3.1-70B), far below the weakest diagnosis score (0.362). Medication action selection alone does not explain the gap: Llama-3.1-8B and Llama-3.1-70B have the highest medication-action accuracies among submitted medication decisions (0.733 and 0.704), but still obtain low medication match scores. The bottleneck is therefore not only choosing whether to start, stop, switch, or adjust treatment; it is selecting the correct drug identity.

\paragraph{Process metrics separate useful interaction from volume.} Across the seven model-level points in Table~\ref{tab:main-results-long-horizon}, information coverage is strongly correlated with decision F1 (Pearson $r=0.865$), moderately anticorrelated with aggregate laboratory waste ratio ($r=-0.496$), and almost uncorrelated with mean laboratory spending ($r=0.103$). Turn count is negatively correlated with coverage ($r=-0.404$) and positively correlated with waste ($r=0.644$). Thus, longer dialogues do not automatically retrieve more useful information; the relevant distinction is whether the model asks clinically targeted questions and orders tests that the ground-truth clinical picture supports.

\end{document}